\documentclass{article}

\usepackage{arxiv}          
\usepackage[utf8]{inputenc} 
\usepackage[T1]{fontenc}    
\usepackage{hyperref}       
\usepackage{url}            
\usepackage{booktabs}       
\usepackage{amsfonts}       
\usepackage{nicefrac}       
\usepackage{microtype}      
\usepackage{graphicx}	      
\usepackage{enumitem}	      
\usepackage{tabularx}       
\usepackage{multirow}       
\usepackage{caption}        
\usepackage{siunitx}        
\usepackage{tikz}           
\usepackage{placeins}       
\usepackage[lofdepth,lotdepth]{subfig}
\usepackage[style=numeric,sorting=none]{biblatex}

\date{} 
\renewcommand{\undertitle}{} 
\renewcommand{\headeright}{} 

\hypersetup{
    colorlinks=true,
    urlcolor=blue,
    pdftitle={On the Importance of Capturing a Sufficient Diversity of Perspective for the Classification of micro-PCBs},
    pdfpagemode=FullScreen,
}

\title{On the Importance of Capturing a Sufficient Diversity of Perspective for the Classification of micro-PCBs}

\author{
  Adam~Byerly\\
  Department of Electronic and Electrical Engineering\\
  Brunel University London\\
  Uxbridge, UB8 3PH UK \\
  Department of Computer Science and Information Systems\\
  Bradley University\\
  Peoria, Il, 61615 USA\\
  \texttt{abyerly@fsmail.bradley.edu} \\
  \And{}
  Tatiana~Kalganova \\
  Department of Electronic and Electrical Engineering\\
  Brunel University London\\
  Uxbridge, UB8 3PH UK \\
  \texttt{tatiana.kalganova@brunel.ac.uk} \\
  \And{}
  Anthony J. Grichnik \\
  Blue Roof Labs \\
  759 N. Main St.\\
  Eureka, IL 61530--1035 USA \\
  Tony.Grichnik@BlueRoofLabs.com \\
}

\begin{document}

\maketitle

\begin{abstract}
We present a dataset consisting of high-resolution images of 13 micro-PCBs captured in various rotations and perspectives relative to the camera, with each sample labeled for PCB type, rotation category, and perspective categories.  We then present the design and results of experimentation on combinations of rotations and perspectives used during training and the resulting impact on test accuracy.  We then show when and how well data augmentation techniques are capable of simulating rotations vs.\ perspectives not present in the training data.  We perform all experiments using CNNs with and without homogeneous vector capsules (HVCs) and investigate and show the capsules' ability to better encode the equivariance of the sub-components of the micro-PCBs.  The results of our experiments lead us to conclude that training a neural network equipped with HVCs, capable of modeling equivariance among sub-components, coupled with training on a diversity of perspectives, achieves the greatest classification accuracy on micro-PCB data.
\end{abstract}


\section{Introduction}\label{sec:introduction}

Prior to the relatively recent successes of deep learning, computer vision research emphasized the encoding of knowledge of geometric models of the objects to be recognized.  In the absence of existing CAD models, a computer vision engineer of this era would need to devise some method of intuiting objects' geometric properties from the data available and manually inputting the information into the target system~\cite{Besl1985}\cite{Chin1986}.  Obviously this is not a scalable solution.  In the mid-1990s researchers turned their attention to the mechanisms employed by biological vision systems and began to focus on the appearance of objects in the 3-D world as projected onto 2 dimensions in order to both classify and estimate the pose of objects.  In 1995, researchers put forth a method for capturing objects of interest from a multitude of positions relative to the camera and, importantly, publicly released their dataset that included 20 objects captured in 90 different poses with 5 positions for the lighting source for a total of 450 images for each of the 20 objects~\cite{Murase1995}.  This dataset would come to be known as the COIL-20 dataset.

Early methods for attempting to classify objects in varying 2-D projections of the 3-D world used classic computer vision methods for identifying edges and contours.  For example, in~\cite{Belongie2001}, the authors measured the similarity of objects after identifying correspondences between images and then using those correspondences to construct an estimated aligning transform.  They were able to achieve an impressive 2.4\% error rate on the COIL-20 dataset when trained from an average of just 4 of the 2-D projections of each 3-D object.

Several years later, researchers generated a larger more difficult dataset, the NORB dataset (NYU Object Recognition Benchmark)~\cite{LeCun2004}.  This dataset included higher resolution images than COIL-20 and contained not just variations in the objects' rotations, but also their perspectives relative to the camera.

Also, worthy of note is the Multi-PIE dataset~\cite{Gross2008}, which is composed, not of objects, but of faces, from 337 different subjects from 15 different viewpoints and having 6 different expressions, for the purposes of facial recognition.  Recognizing a specific face among a group of faces, as opposed to recognizing a four-legged animal among cars and airplanes, is a significant challenge because, assuming a lack of deformity, all faces are made up of the same sub-parts in similar locations (i.e. 2 eyes above a nose, which is above a mouth).

In the past several years, our vocabulary in the field has matured and turned to describing the relationships between features of 2-D projections of 3-D objects as being \textit{equivariant}, a term borrowed from the representation theory of finite groups.  \textit{Translation equivariance} in convolutional neural networks (CNNs), for example, is demonstrated when shifting an image that is fed into a network produces the same result as shifting the output feature maps of the original image in the same direction.  For computer vision applications, the ideal would be for all possible 2-D projections of a 3-D object to possess a calculable or learnable representation of their equivariant properties.  One principal area of research involved in achieving better equivariance involves novel neural network architectural elements designed to facilitate it~\cite{Cohen2016}\cite{Lu2017}\cite{Liu2018}\cite{Lenc2019}.

\textit{Invariance} is the special case of equivariance wherein the transformation is a null transformation.  CNNs are especially adept at successfully identifying features in a translationally invariant manner.  In~\cite{Hinton2011}, the authors critiqued this as being an undesirable property.  They noted that individual features of a larger image are not translationally invariant to one another, but rather possess a specific non-null translational equivariance relative to one another (such as in a person's face, the position of the eyes relative to the nose and the nose relative to the mouth).  They also noted that the quite successful computer vision system SIFT~\cite{Lowe1999}, used hand-engineered feature detectors that were 128-dimensional vectors called keypoint descriptors.  In that work, they put forth a novel neural network architectural element they called capsules that would act much like the scalar-valued neuron, but be vector valued like the keypoint descriptors of SIFT.\@ Using back-propagation, the capsules could then, if exposed to the same objects from different 2-D projections, implicitly learn a vector of values for the equivariance of the feature that capsule detected.  In~\cite{Hinton2018}, the authors applied capsules to the smallNORB dataset (which is a subset of NORB) and achieved a new state-of-the-art test error rate.

Printed Circuit Boards (PCBs) are like the human face in that, for a given PCB, the individual elements (such as capacitors, resistors, and integrated circuits (ICs)) are not present invariantly relative to one another, but rather at very specific locations relative to one another.  Compared to the human face, even on small PCBs, there are a greater number of features with greater similarity to one another (some modern small surface-mounted capacitors and resistors are nearly indistinguishable from one another, whereas eyes, noses, and mouths are quite distinctive from one another).

In~\cite{Pramerdorfer2015b}, the authors put forth a small, medium-resolution dataset of a variety of PCBs and applied SIFT and other similar methods to the task of locating and categorizing the components on the board (i.e.\ identifying resistors vs.\ capacitors vs.\ ICs).  This dataset has a very small number of images and annotates the category and locations of the components on the PCBs.  Likewise work by~\cite{Lu2020},~\cite{Pramerdorfer2015}, and~\cite{Mahalingam2019} introduced similar datasets at higher resolutions.  The unifying characteristic of these datasets is that they all captured relatively few images of any one PCB, all from a completely neutral perspective (i.e.\ the camera directly above the PCB), and that the intended purpose was for classification and categorization of the many sub-components on the PCBs.  Other work has further focused on detecting specifically only surface mounted components~\cite{Li2013}, through-hole components~\cite{Herchenbach2013}, on-board printed text\cite{Li2014}, or manufacturing defects~\cite{Huang2019}.  Again, all of these were focused on analysis of the individual sub-components of the PCBs.  The motivation in all of these cases is to facilitate automated analysis to optimize recycling practices.

There are indeed a truly massive number of different PCBs that do or could exist.  As such, the analysis of sub-components does make sense for the general case---so that knowledge of the specific make and model of a PCB is not a prerequisite to knowing the sub-components used in it.  However, if the distribution of PCB makes and models follows a Pareto distribution (a conjecture we do not intend to demonstrate), then it stands to reason that a small number of PCB makes and models could be far more common.  Assuming this hypothesis, a candidate for the types of PCBs that would be more common are those that are general-purpose, affordable, and small.  We refer to this class of PCBs as micro-PCBs.  In those cases, accurately classifying the make and model of the PCB, rather than attempting to locate and categorize every sub-component, could, through a bill of materials, accurately identify \textit{all} of the sub-components on such PCBs.  In this case, robustness to environmental factors during image acquisition through a wider range of rotations and perspectives becomes an enabling factor for implementations.

Our contribution is as follows:
\begin{enumerate}
  \item We present a dataset consisting of high-resolution images of 13 micro-PCBs captured in various rotations and perspectives relative to the camera, with each sample labeled for PCB type, rotation category, and perspective category.  This dataset is unique relative to the other available PCB datasets (such as~\cite{Pramerdorfer2015b},~\cite{Lu2020},~\cite{Pramerdorfer2015}, and~\cite{Mahalingam2019}) in that the micro-PCBs in our dataset are (a) readily available and inexpensive models, (b) captured many more times each, and (c) captured from a variety of labeled rotations and perspectives.
  \item We conduct experiments on the dataset using CNNs with and without capsules and investigate and show the capsules' ability to better encode the equivariance of the sub-components of the micro-PCBs.
  \item We conduct experiments on the dataset investigating the ability of data augmentation techniques to improve classification accuracy on novel rotations and perspectives of the micro-PCBs when using CNNs with and without capsules.
  \item The results of our experiments show that, as rotations can be more effectively and accurately augmented, image acquisition efforts should prioritize the capturing of a diversity of perspectives representative of the perspectives for which accurate inference is important.
\end{enumerate}

\section{Image Acquisition}\label{sec:image_acquisition}

\begin{figure}[!ht]
  \centering
  \setlength\tabcolsep{1pt}
  \begin{tabular}{@{}cccc@{}}
    \includegraphics[width=0.19\textwidth]{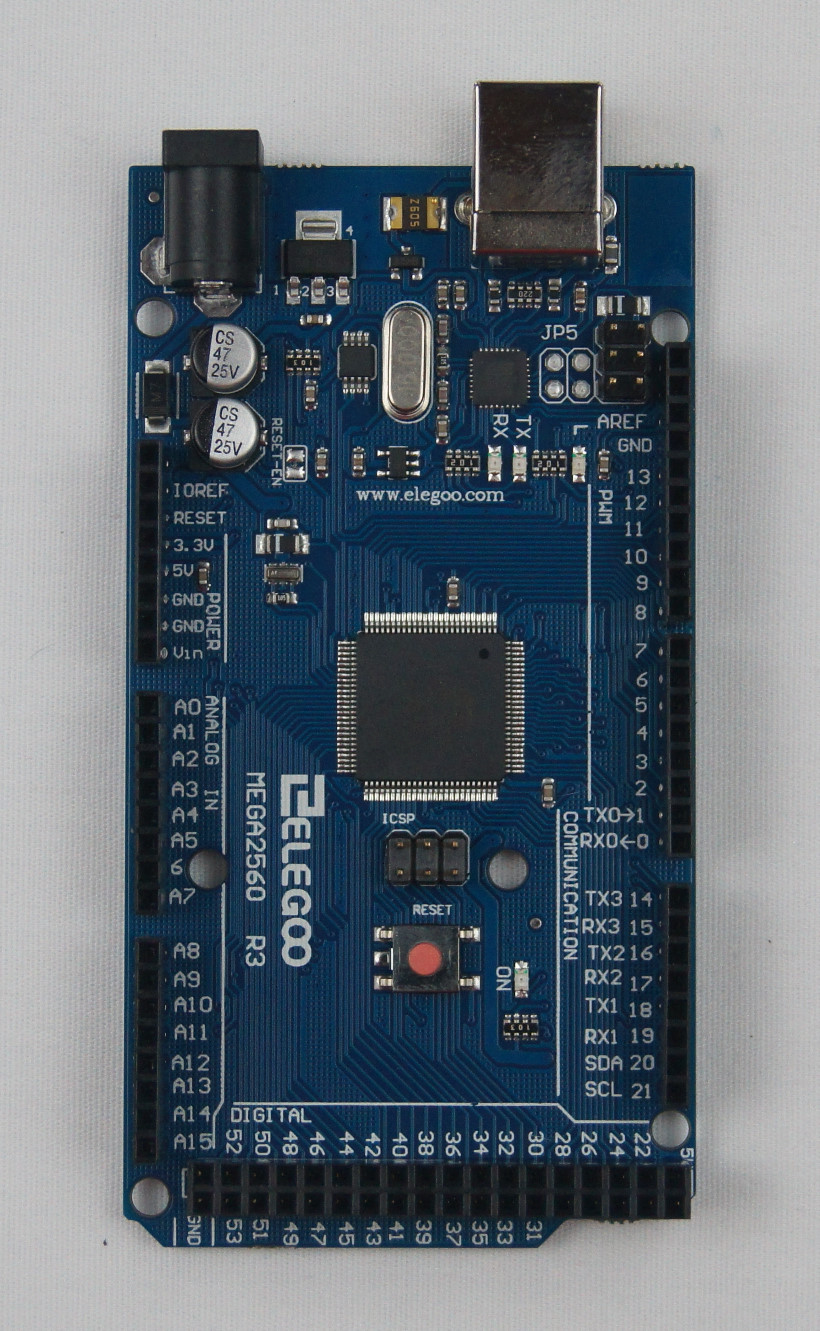} &
    \includegraphics[width=0.19\textwidth]{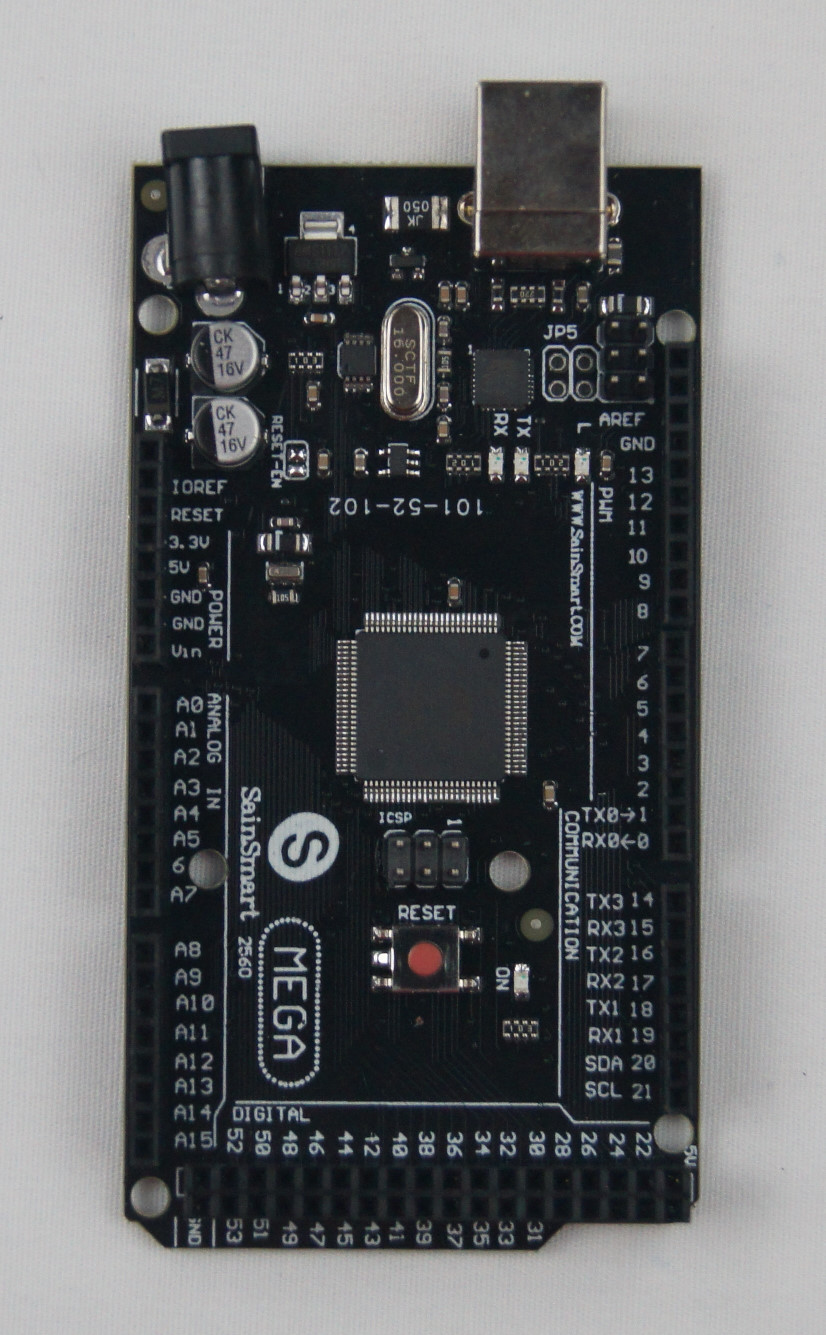} &
    \includegraphics[width=0.19\textwidth]{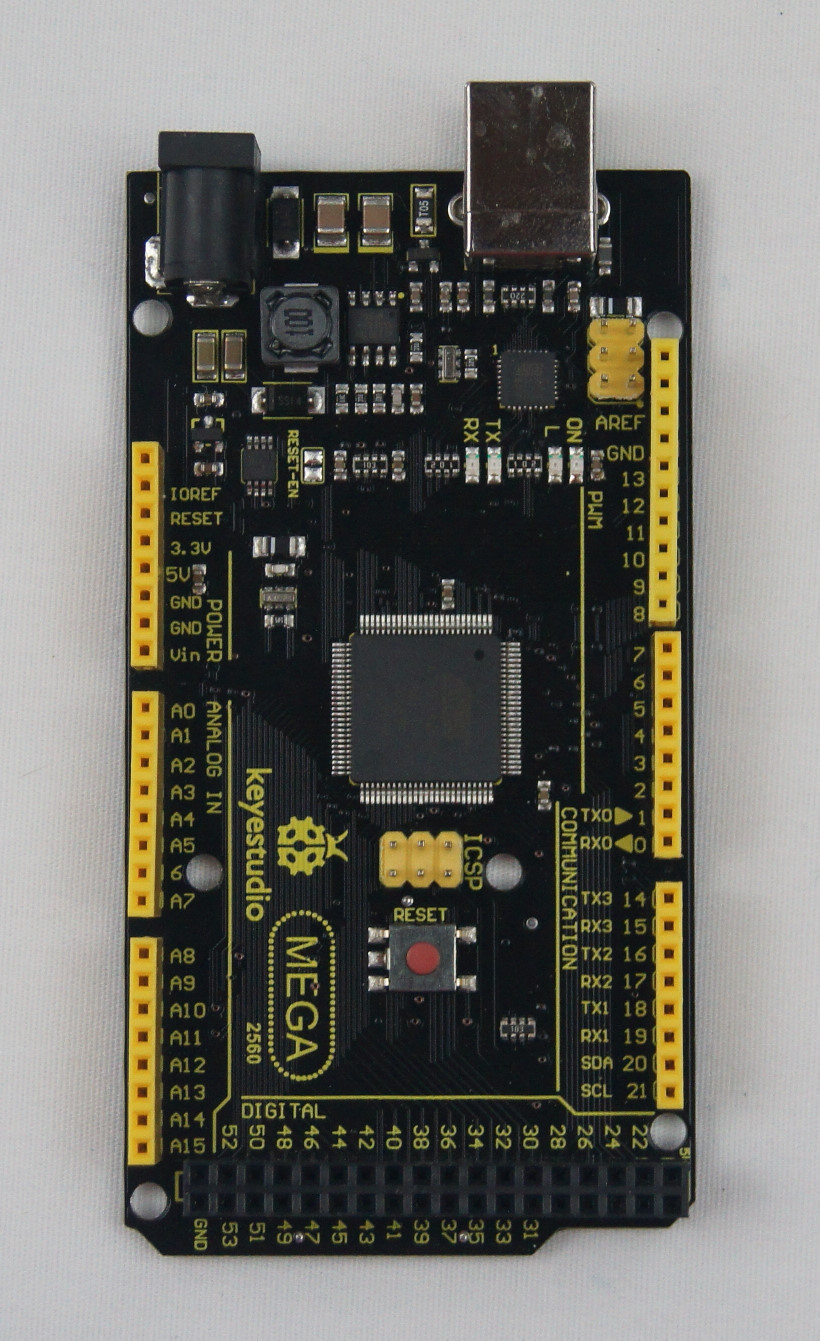} &
    \includegraphics[width=0.19\textwidth]{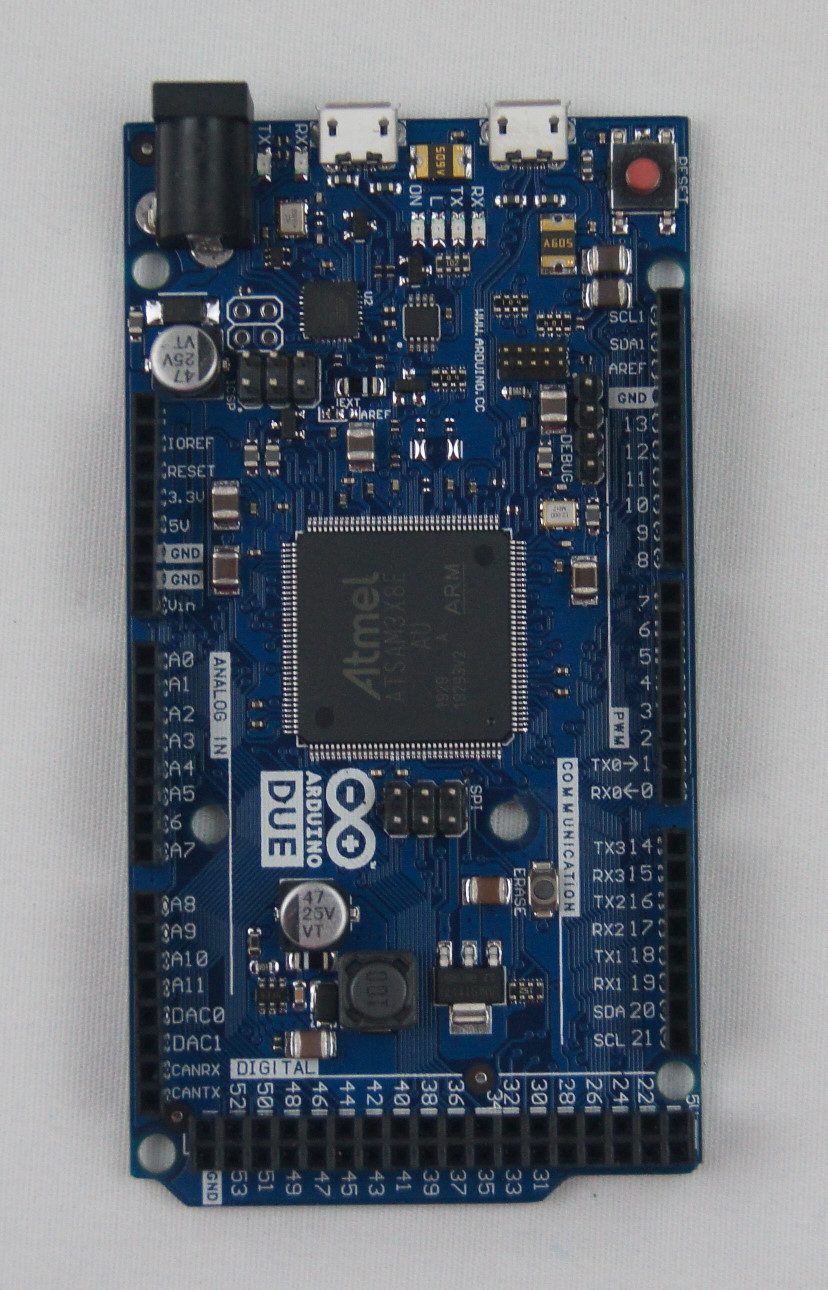} \\
    (a) & (b) & (c) & (d) \\
  \end{tabular}
  \begin{tabular}{@{}cccc@{}}
    \includegraphics[width=0.19\textwidth]{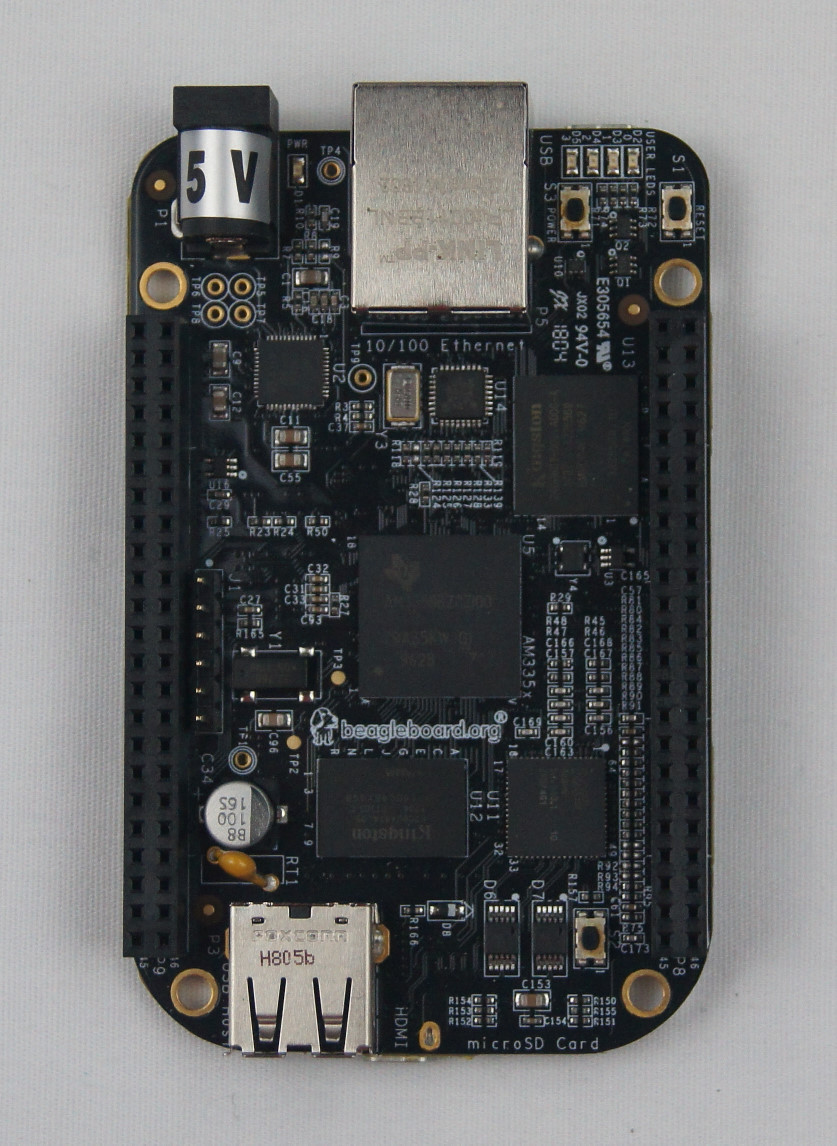} &
    \includegraphics[width=0.19\textwidth]{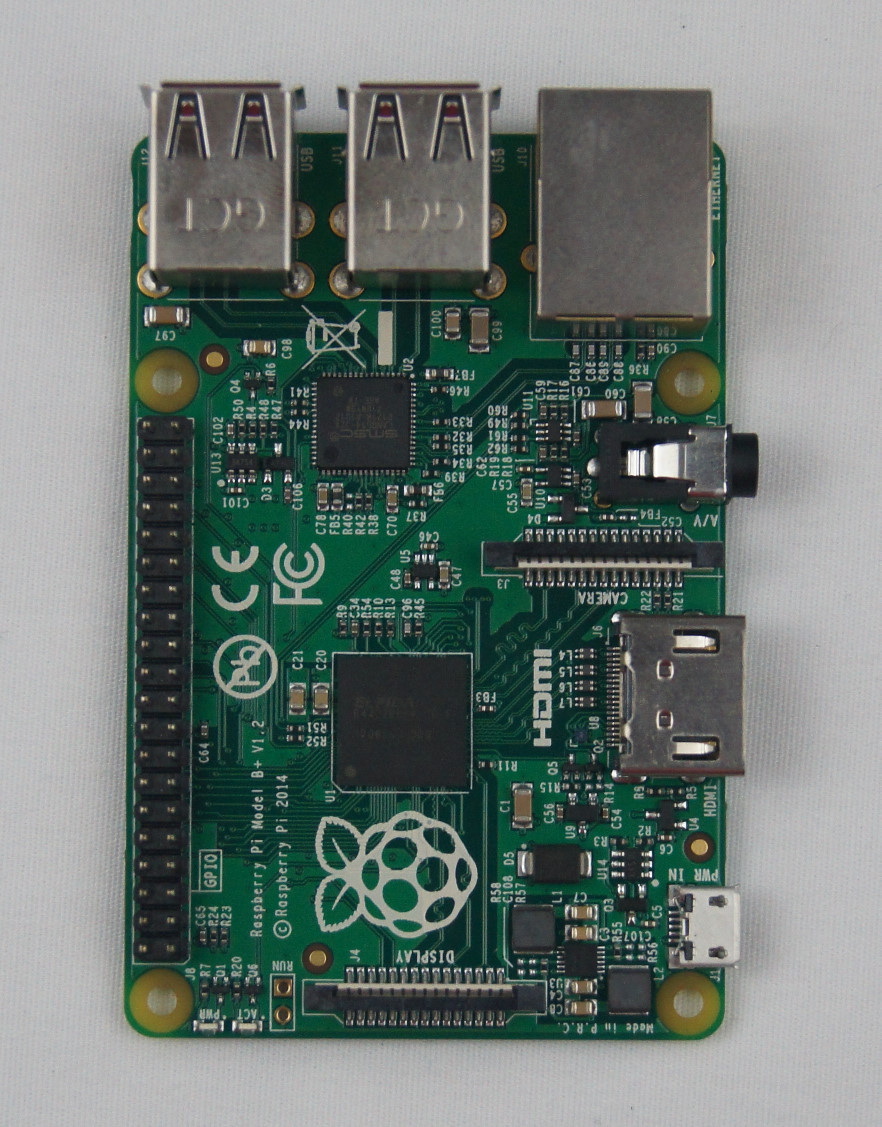} &
    \includegraphics[width=0.19\textwidth]{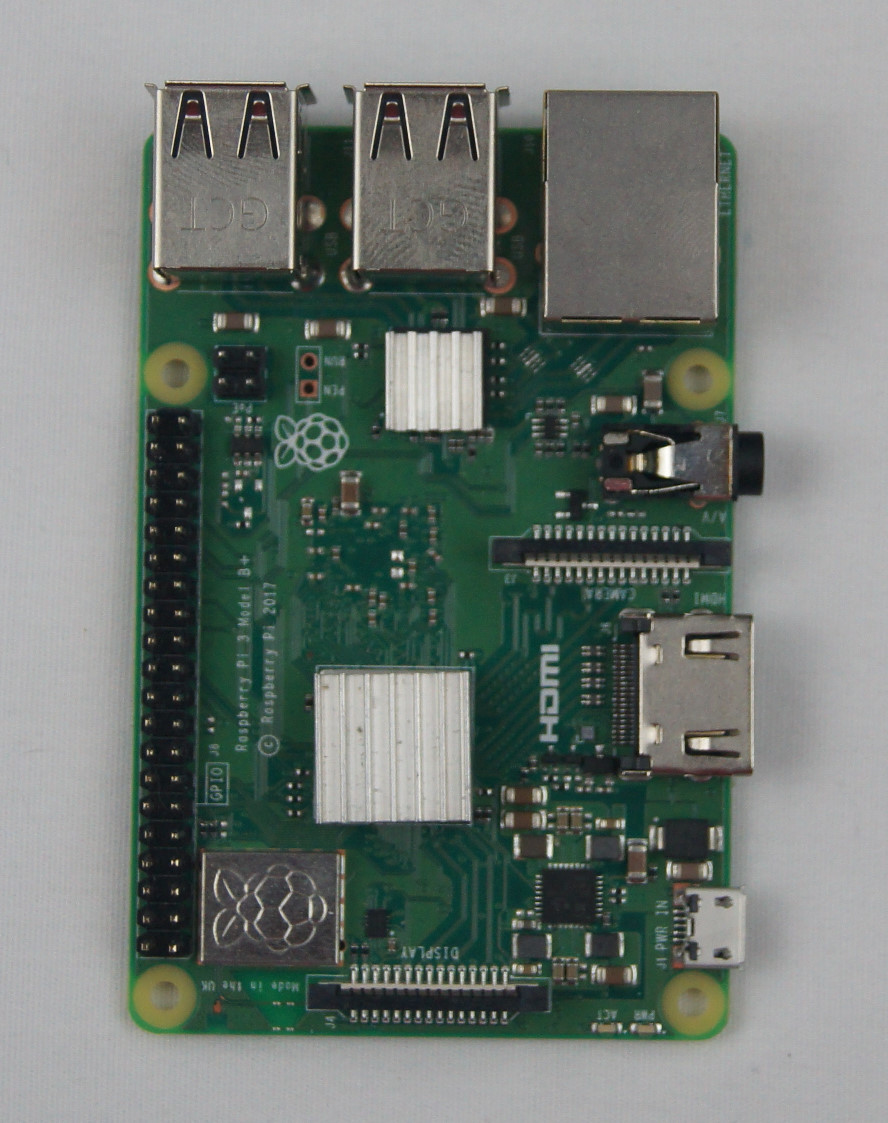} &
    \includegraphics[width=0.19\textwidth]{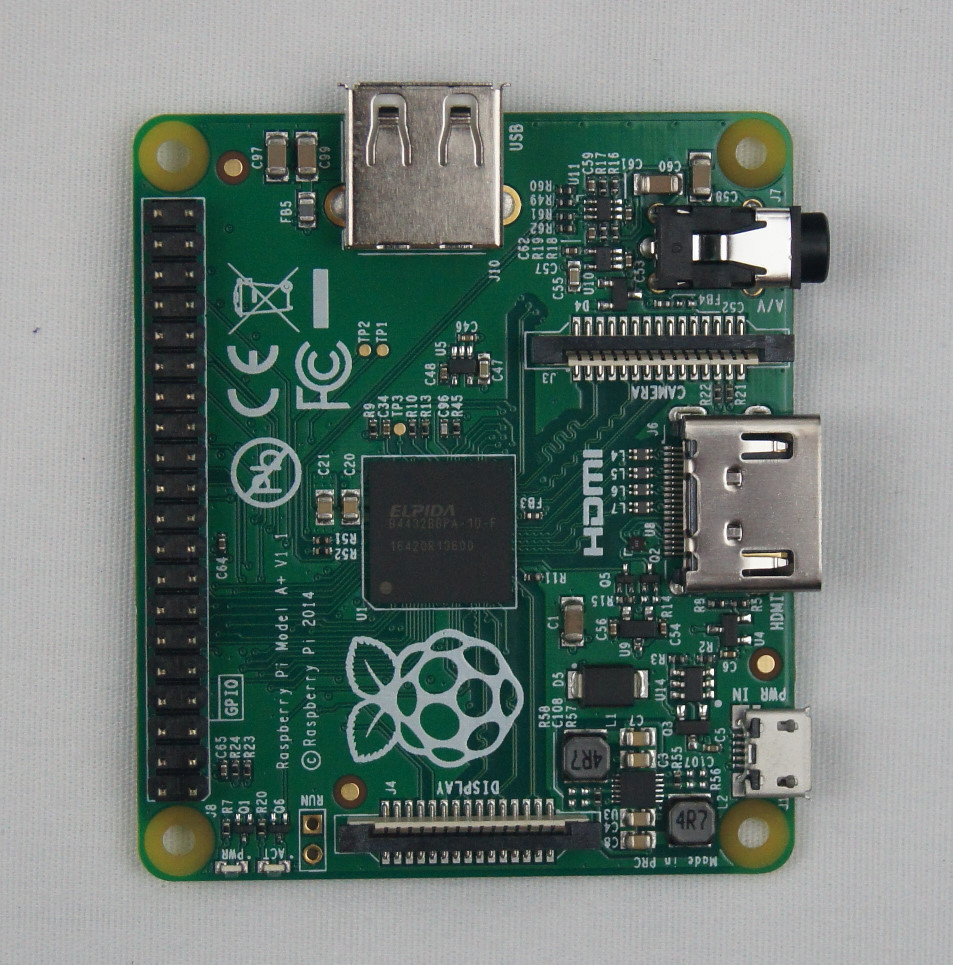} \\
    (e) & (f) & (g) & (h) \\
  \end{tabular}
  \begin{tabular}{@{}ccccc@{}}
    \includegraphics[width=0.19\textwidth]{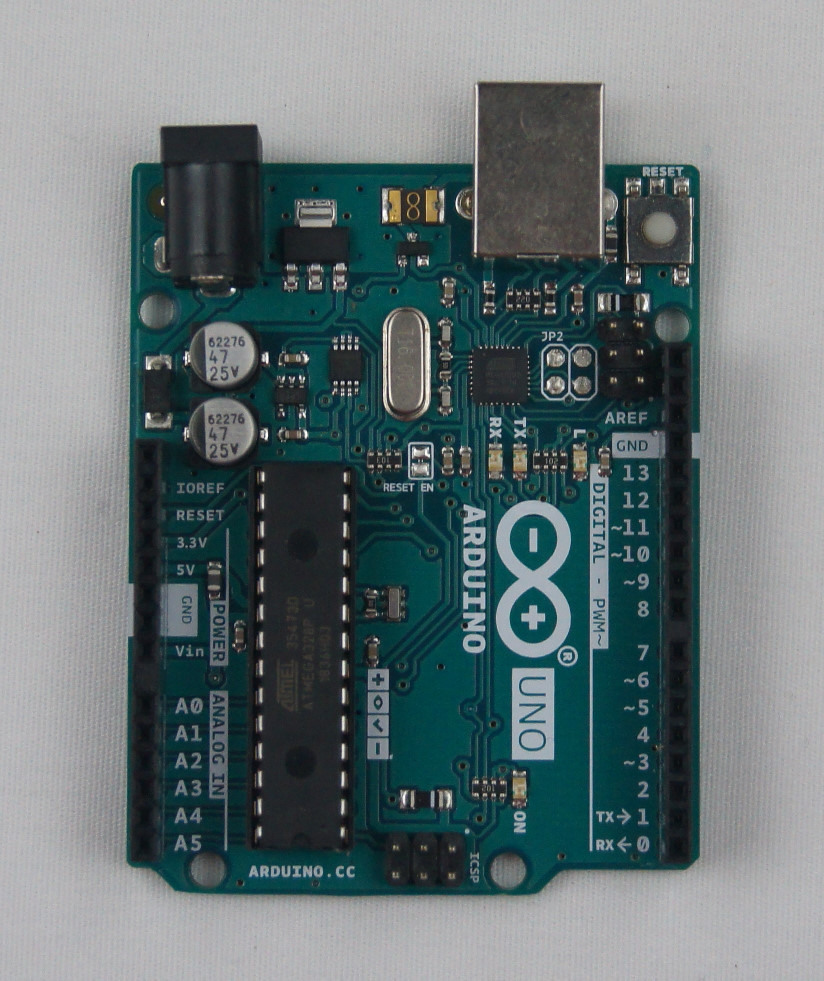} &
    \includegraphics[width=0.19\textwidth]{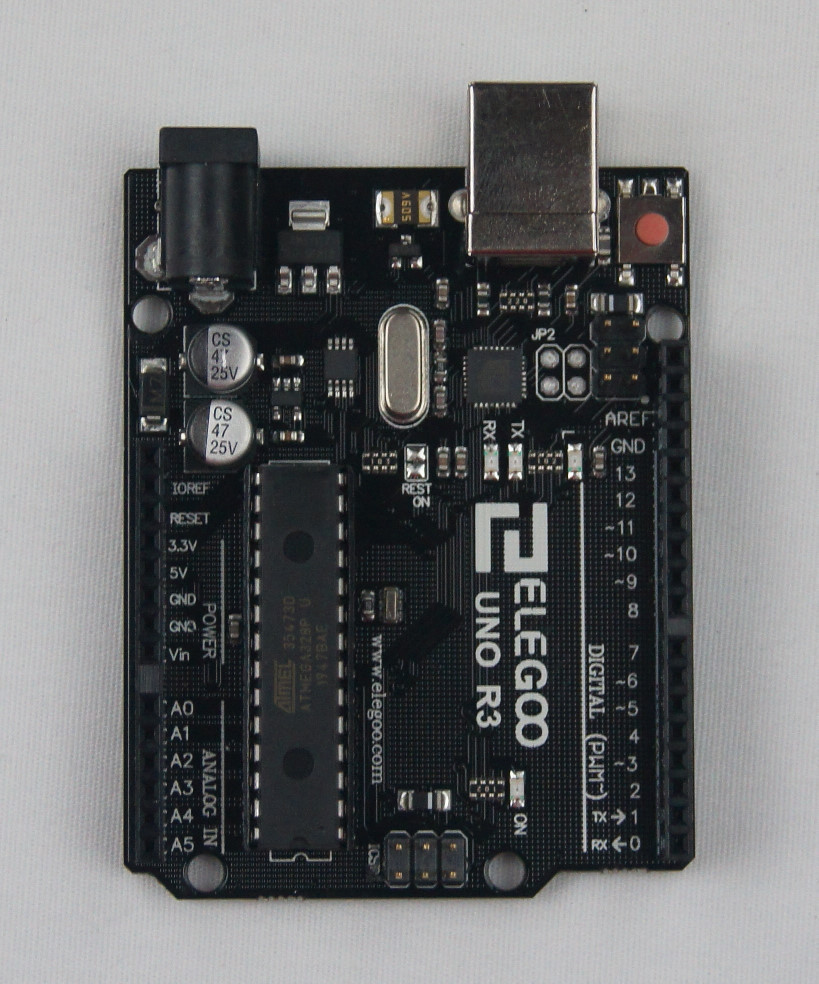} &
    \includegraphics[width=0.19\textwidth]{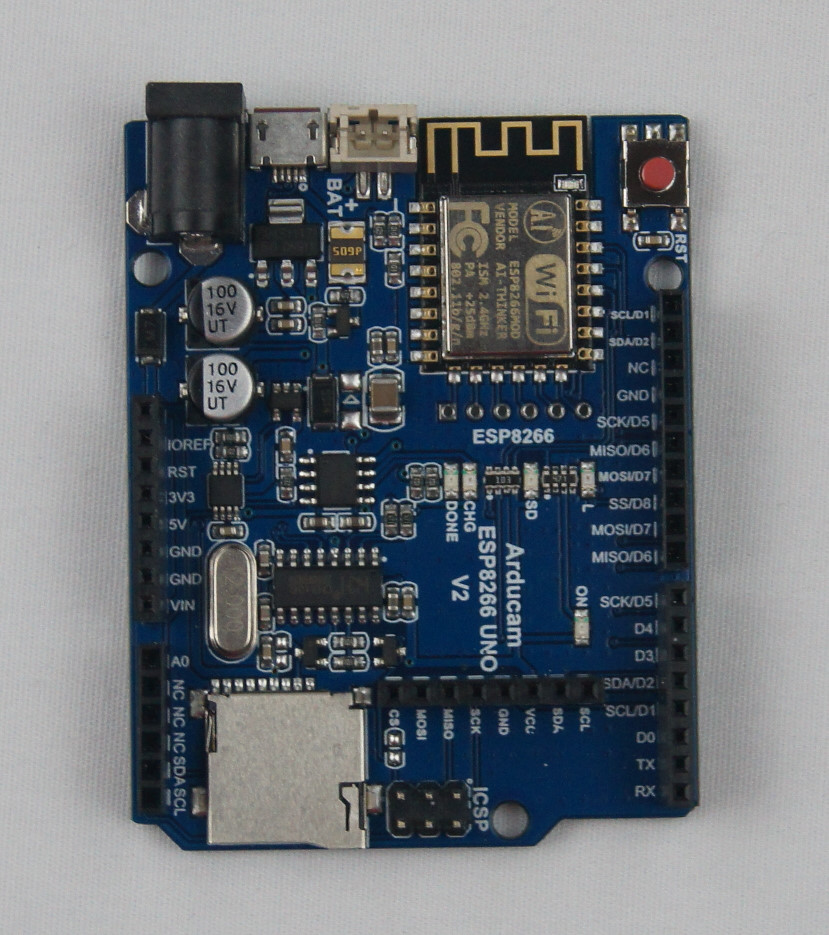} &
    \includegraphics[width=0.19\textwidth]{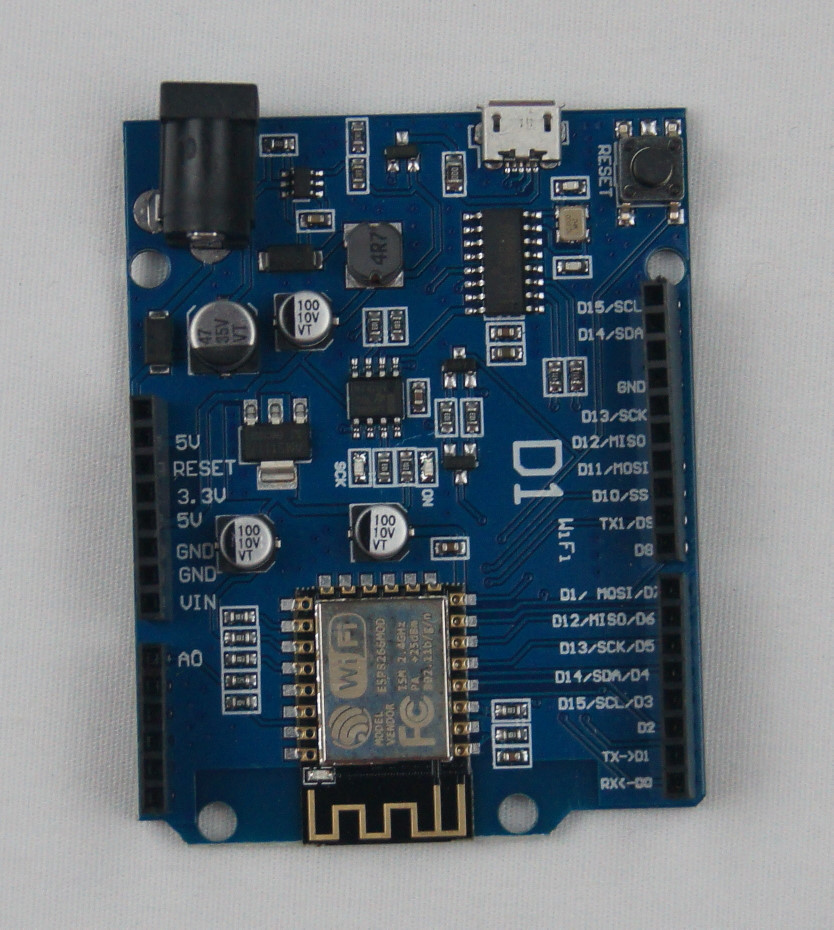} &
    \includegraphics[width=0.19\textwidth]{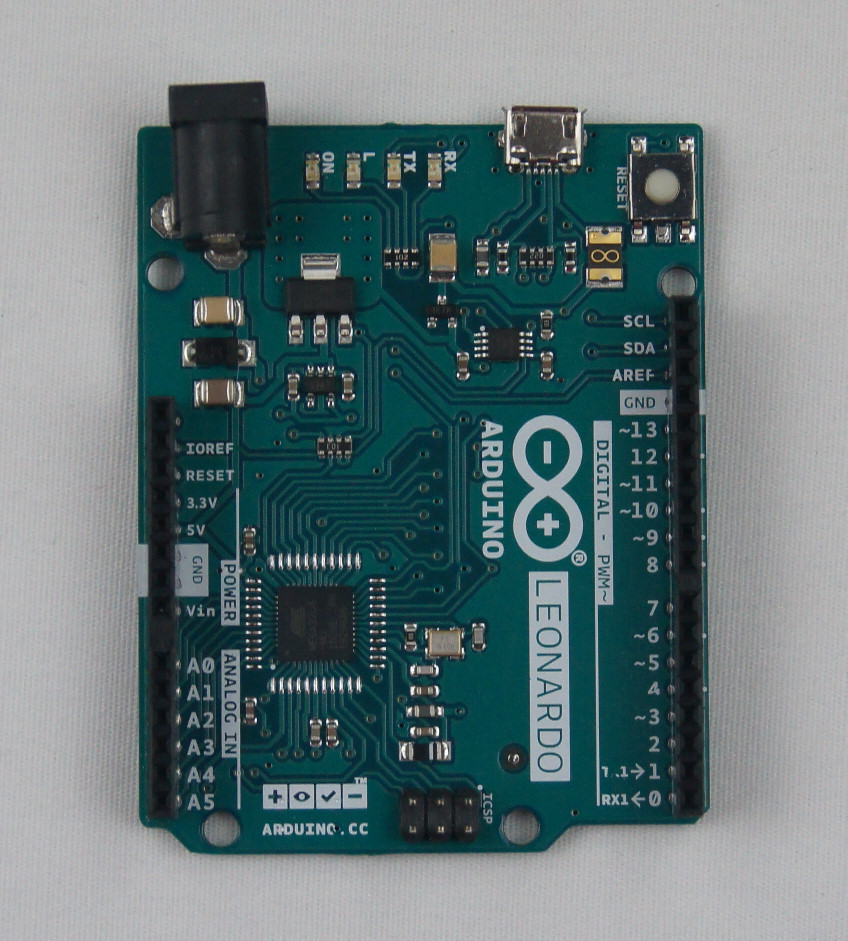} \\
    (i) & (j) & (k) & (l) & (m) \\
  \end{tabular}
  \caption{The 13 micro-PCBs for which images were acquired. (a-c) Arduino Mega 2560 from 3 different manufacturers. (d) Arduino Due. (e) Beaglebone Black. (f) Raspberry Pi 1 B+. (g) Raspberry Pi 3 B+. (h) Raspberry Pi A+. (i-j) Arduino Uno from 2 different manufacturers. (k) Arduino Uno WiFi Shield. (l) Arduino Uno Camera Shield. (m) Arduino Leonardo. }\label{fig:board_types}
\end{figure}

We captured a total of 8,125 images of the 13 micro-PCBs (see \autoref{fig:board_types}) in our dataset using a Sony SLT-A35, 16 Megapixel DSLR Camera, in Advanced Auto mode and using a polarization filter.  After cropping the excess area around the micro-PCBs in each image, the average size of all images is 1949\(\times\)2126 (width\(\times\)height).

\begin{figure}[!ht]
  \centering
  \includegraphics[width=0.66\textwidth]{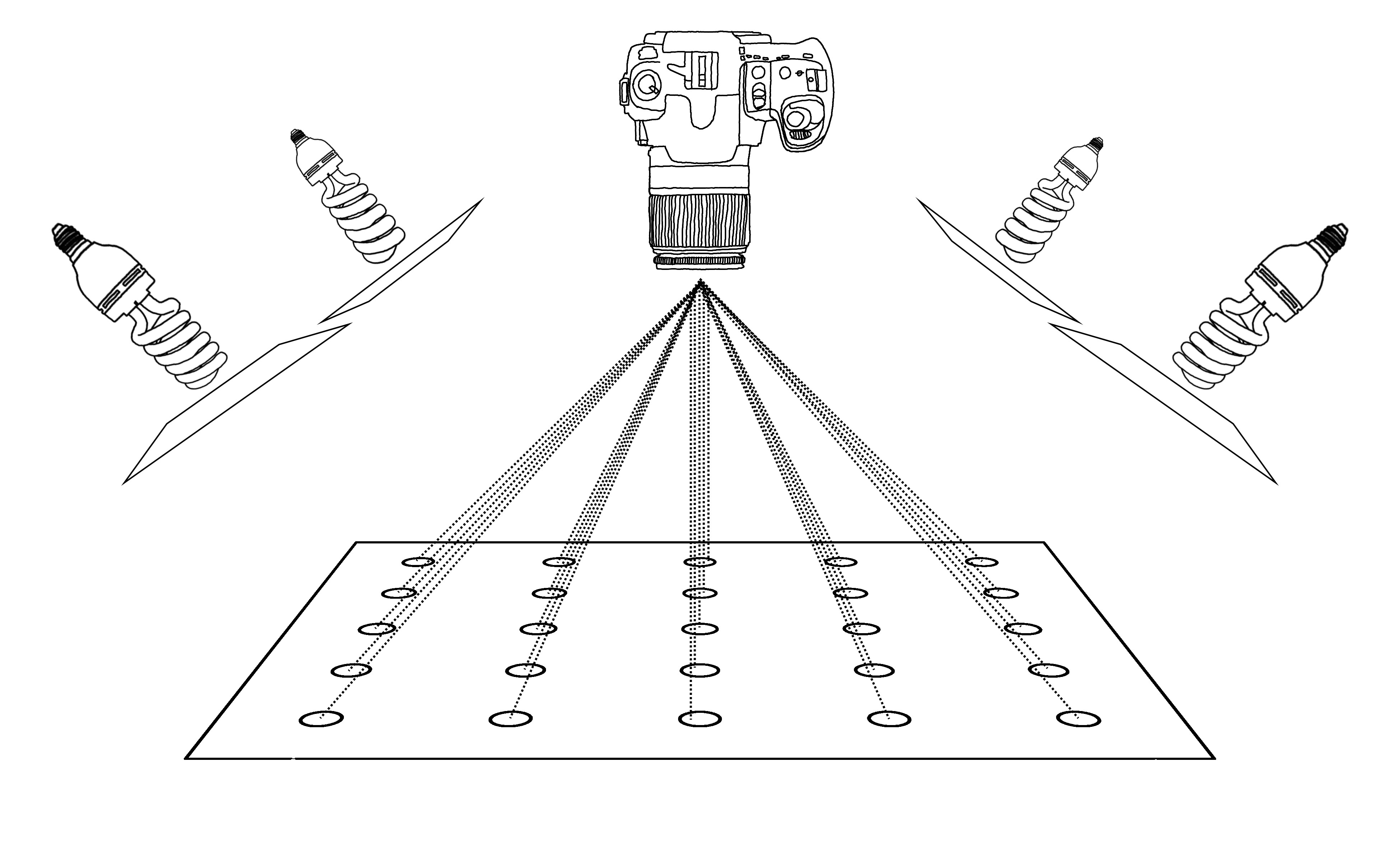}
  \caption{A depiction of the image acquisition environment.  Each micro-PCB were captured in 25 different positions in a \(5\times5\) grid on the capture surface such that each position was 6 inches from its horizontal and vertical neighbors.  The camera was positioned 16 inches directly above the central position in the grid.  The light sources were placed 16 inches outside of the grid in the four corners of the grid and 32 inches above the capture surface, each with diffusion material directly in front of the bulbs.}\label{fig:lighting_rig}
\end{figure}

\begin{figure}[!ht]
  \centering
  \begin{minipage}{.7\textwidth}
    \begin{tabularx}{\textwidth}{@{}cXc@{}}
      \includegraphics[width=1.9in]{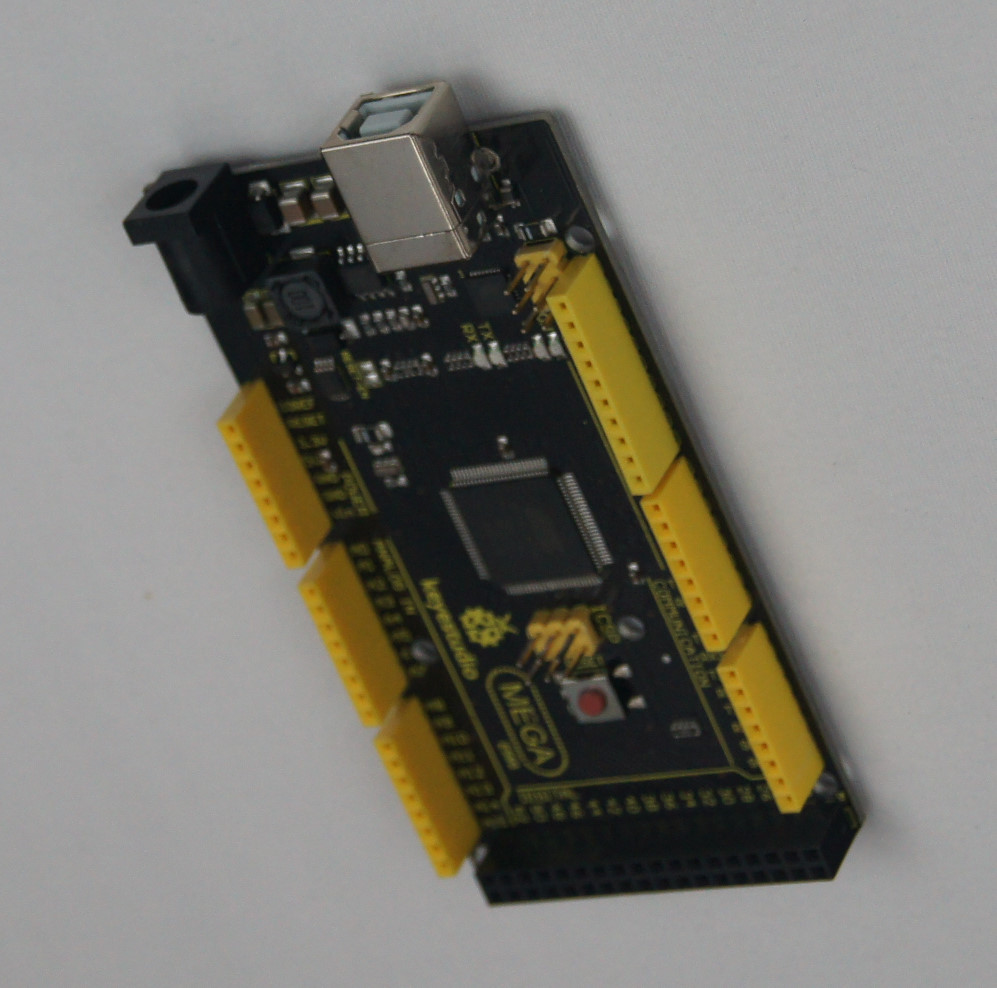} &
      &
      \includegraphics[width=1.5in]{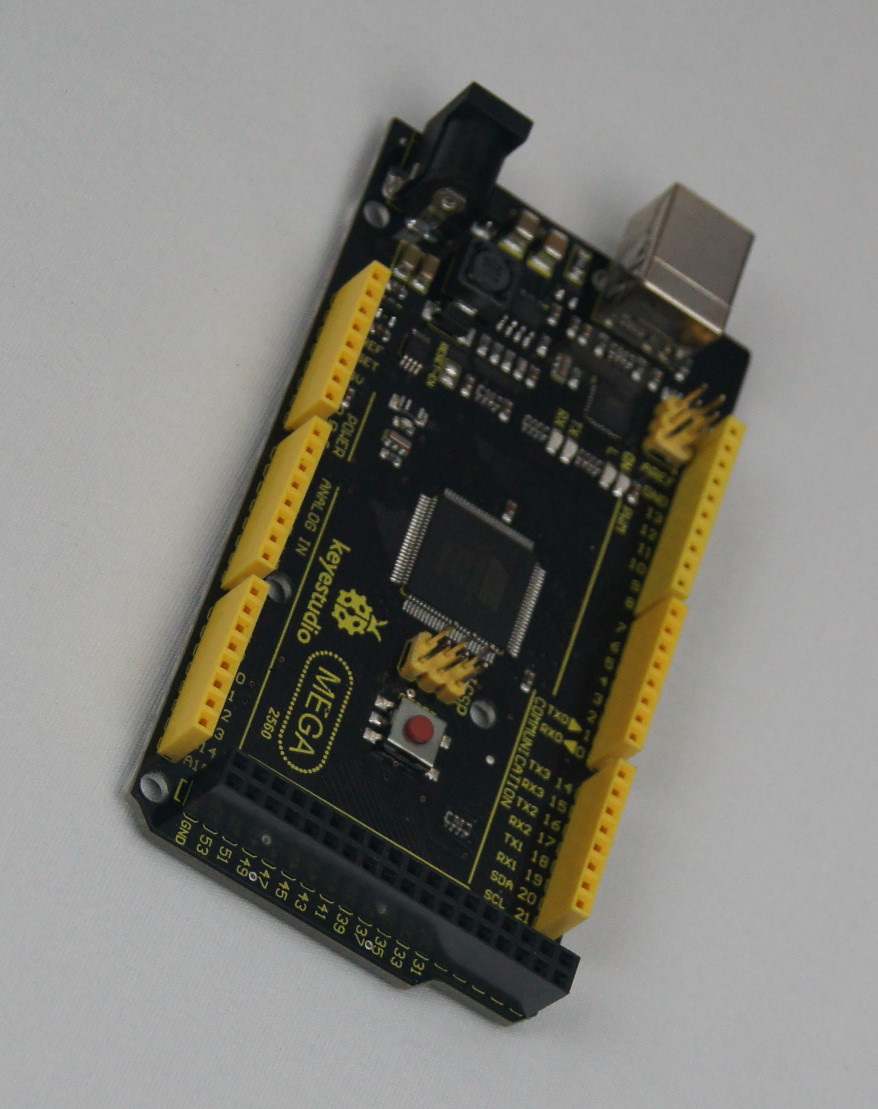} \\
    \end{tabularx}
    \caption{Examples of one of the manufacturer's Arduino Mega 2560 captured from two different extreme perspectives on the capture surface relative to the camera.}\label{fig:board_extremes}
  \end{minipage}
\end{figure}

\begin{figure}[!ht]
  \centering
  \setlength\tabcolsep{1pt}
  \begin{tabular}{@{}ccccc@{}}
    \includegraphics[width=0.19\textwidth]{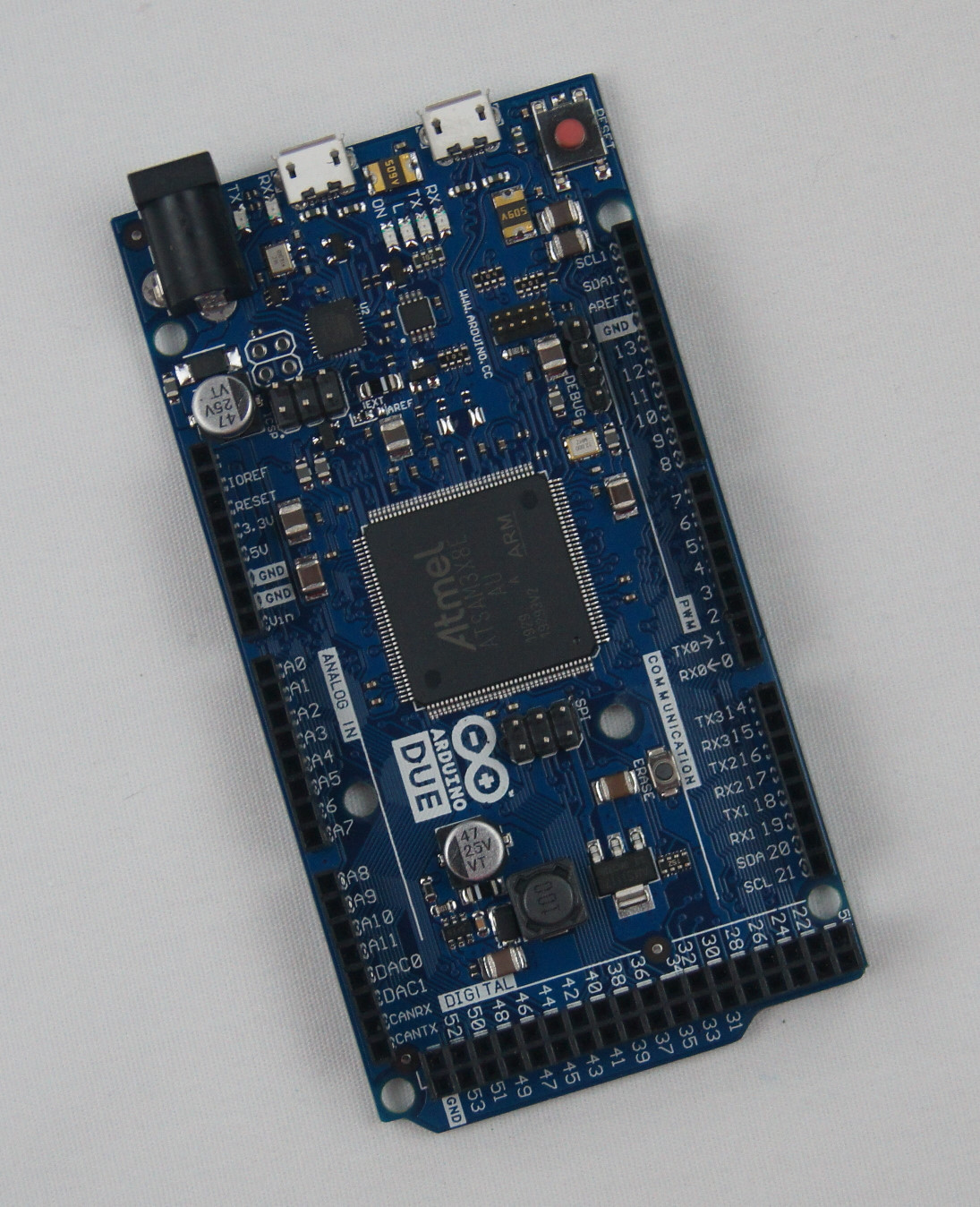} &
    \includegraphics[width=0.19\textwidth]{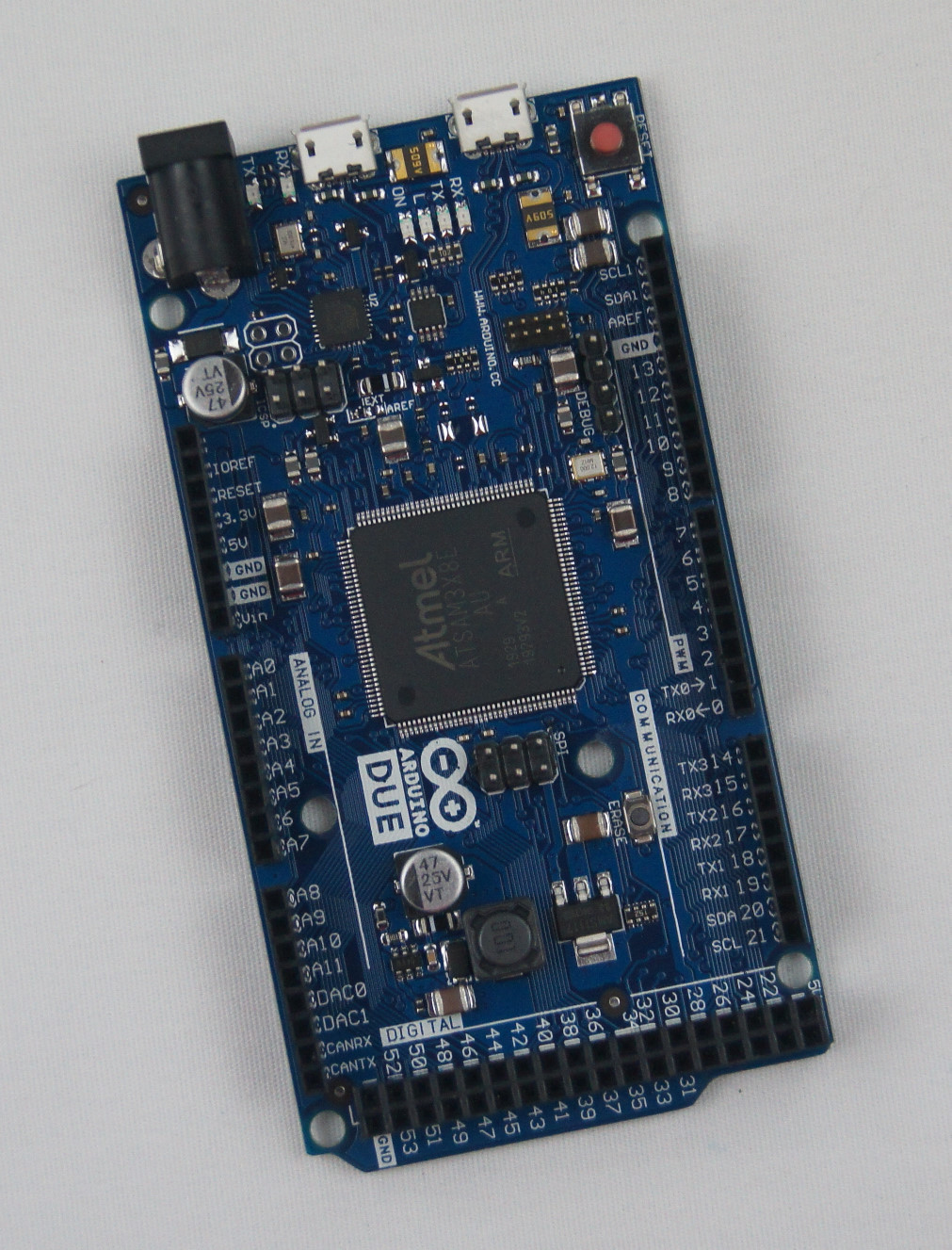} &
    \includegraphics[width=0.19\textwidth]{images/DCCC1.jpg} &
    \includegraphics[width=0.19\textwidth]{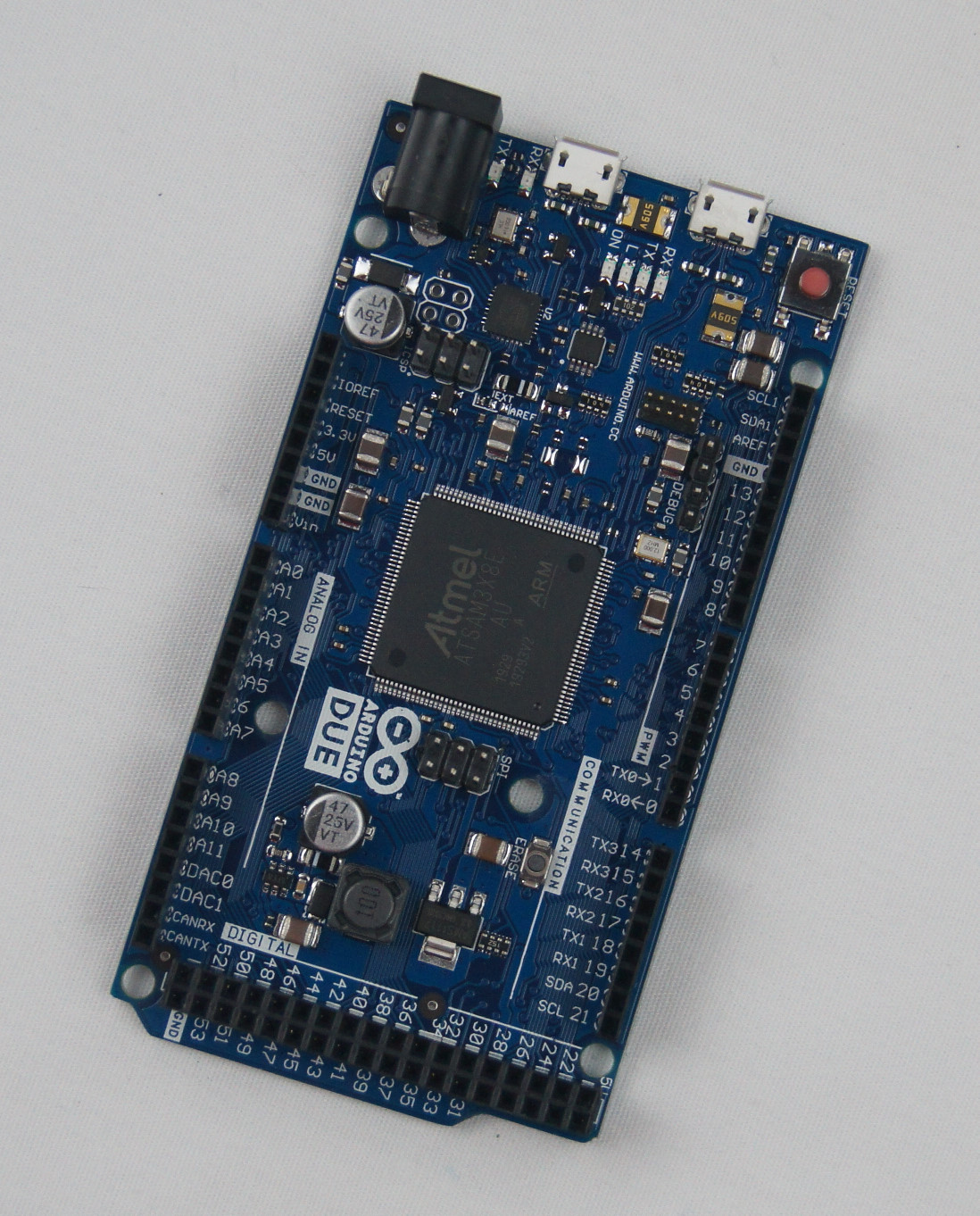} &
    \includegraphics[width=0.19\textwidth]{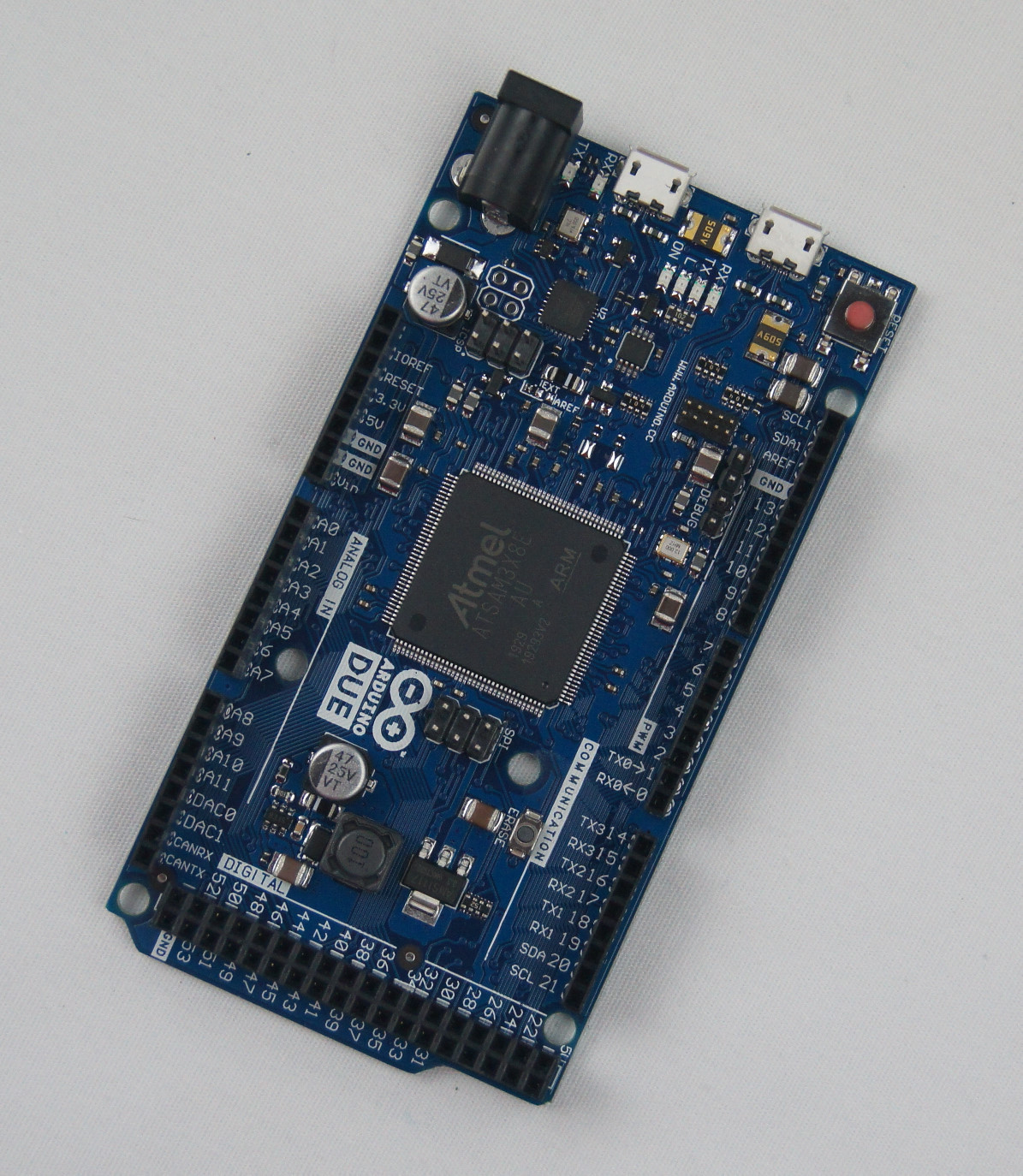} \\
  \end{tabular}
  \caption{Examples of the Arduino Due captured in the neutral perspective position from the five different rotations each micro-PCB was captured from.}\label{fig:board_angles}
\end{figure}

\begin{figure}[!ht]
  \centering
  \hfill
  \subfloat[Printing on the capacitors present on one manufacturer's Arduino Uno that differs between the micro-PCB coded for training and the one coded for testing.]{
    \begin{tabular}{@{}cc@{}}
      \includegraphics[width=0.19\textwidth]{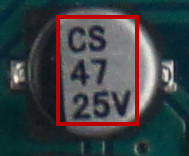} &
      \includegraphics[width=0.19\textwidth]{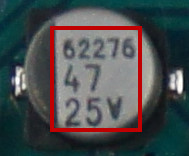} \\
    \end{tabular}\label{fig:inconsequential_dif_a}}
  \hfill
  \subfloat[Printing on the USB receiver present on the Beaglebone Black that differs between the micro-PCB coded for training and the one coded for testing.]{
    \begin{tabular}{@{}cc@{}}
      \includegraphics[width=0.19\textwidth]{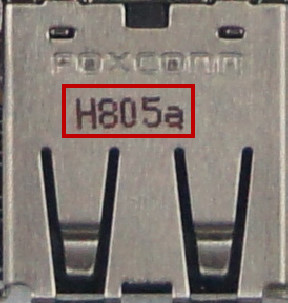} &
      \includegraphics[width=0.19\textwidth]{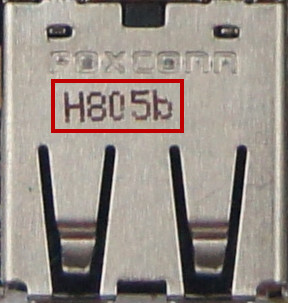} \\
    \end{tabular}\label{fig:inconsequential_dif_b}}
  \hfill{} 
  \caption{Examples of inconsequential printing differences present on the micro-PCBs that were coded for training vs.\ those coded for testing.}\label{fig:inconsequential_dif}
\end{figure}

The micro-PCBs were captured in 25 different positions relative to the camera (see \autoref{fig:lighting_rig}) under ideal lighting conditions using (4) 85-Watt CFL Full Spectrum 5500K color bulbs each producing approximately 5,000 lumens.  In each position, each micro-PCB was captured in 5 different rotations (see \autoref{fig:board_extremes} and \autoref{fig:board_angles}).  This creates 125 unique orientations of each micro-PCB relative to the camera.  Each unique orientation was captured 4 times and coded for training and then another micro-PCB of the same make and model was captured once and coded for testing.  Thus, no micro-PCB that is used in training is the same that is used in testing.  Although the micro-PCBs coded for training are nearly identical to those coded for testing, very subtle differences exist in some cases (see \autoref{fig:inconsequential_dif}).  In total, each micro-PCB in the dataset has 500 training images and 125 test images, creating an overall train/test split of 6,500/1,625.

The micro-PCBs being placed in 25 different positions in the capture surface results in the creation of 25 unique perspectives of each micro-PCB relative to the camera.  We refer to the position directly under the camera as the neutral perspective, the 8 positions directly adjacent to the neutral position as ``near'' perspectives and the outer 16 positions as ``far'' perspectives.  To fully distinguish the 25 perspectives, when looking down from the camera's position to the capture surface, we refer to those that are to the left or above the camera as ``negative'' and those that are to the right or below the camera as ``positive''.  In each perspective, each micro-PCB was rotated across 5 rotations manually without attempting to place them in any exact angle.  Instead, we placed each micro-PCB (1) straight, which we refer to as the neutral rotation, (2--3) rotated slightly to the left and right, which we refer to as the left shallow and right shallow rotations respectively, and (4--5) rotated further to the left and right, which we refer to as the left wide and right wide rotations respectively.  

The goal of computer vision applications is to learn true representations of the objects and not merely to memorize pixel intensities and locations.  The manual placement of the micro-PCBs as well as placing them in rotation without measurement helps to facilitate learning the true representation by introducing small perturbations in the positions of the micro-PCBs.

After image acquisition, we used an edge detection algorithm to detect the left and right edges of each image.  Using the left edge, we computed the angle of each micro-PCB relative to an ideal neutral.\@ See \autoref{tab:angles_of_rotated_images} for statistics regarding these angles.  We then measured the distance of the left edge to the right edge at the bottom and top of each image and created a ratio between the two distances.  This ratio is representative of the true perspective along the \(y\)-axis.  See \autoref{tab:ratios_of_perspectives} for statistics regarding these ratios.  The presence of various connectors on the top edge of the micro-PCBs made algorithmically determining an accurate top edge of the micro-PCBs impossible, so we do not present ratios to be representative of the true perspective along the \(x\)-axis.  However, a reasonable estimate can be calculated using the corresponding ratio for the \(y\)-axis multiplied by the ratio of an image's width to its height.

\begin{table}[!ht]
  \centering
  \begin{minipage}{.7\textwidth}
    \caption{Angles of Rotated Images}
    \begin{tabularx}{\textwidth}{@{}Xrrrr@{}}
      \toprule
        Position & Min & Mean & S.D. & Max \\
        \midrule
        Left Wide	    & 10.43	& 21.31	& 4.50481	& 32.78602043 \\
        Left Shallow	&  4.23	& 12.39	& 3.59036	& 23.72892221 \\
        Neutral     	&     0	& 2.475	& 2.04823	& 12.76094982 \\
        Right Shallow &  0.16	& 14.73	& 4.46218	& 31.52155152 \\
        Right Wide  	&  0.76	& 24.31	& 5.27139	& 42.84832537 \\
        \bottomrule
    \end{tabularx}\\\label{tab:angles_of_rotated_images}
    \captionsetup{justification=justified,singlelinecheck=false}
    \caption*{Values are in degrees and represent the absolute value of the deviation from a true neutral rotation. (S.D. is standard deviation.)}
  \end{minipage}
\end{table}

\begin{table}[!ht]
  \centering
  \begin{minipage}{.7\textwidth}
    \caption{Ratios of micro-PCB Width Differences}
    \begin{tabularx}{\textwidth}{@{}Xrrrr@{}}
      \toprule
        Position & Min & Mean & S.D. & Max \\
        \midrule
        Negative Far	  & 0	& 12.71\%	& 6.27130 & 48.18\% \\
        Negative Near	  & 0	&  7.94\%	& 5.05283 & 29.66\% \\
        Neutral         & 0	&  4.40\%	& 3.50910 & 20.90\% \\
        Positive Near   & 0	&  7.09\%	& 3.59487 & 26.76\% \\
        Positive Far  	& 0 & 11.45\%	& 4.36561 & 23.13\% \\
        \bottomrule
    \end{tabularx}\\\label{tab:ratios_of_perspectives}
    \captionsetup{justification=justified,singlelinecheck=false}
    \caption*{Edge detection algorithms were used to identify the left and right edges of the micro-PCBs.  The values here are absolute values of ratios of the width between the detected edges at the bottom and top of the image.}
  \end{minipage}
\end{table}

\section{Experimental Design and Results}\label{sec:experimental_design}

In~\cite{Byerly2019}, the authors experimented with a simple monolithic CNN.\@ In their experiments, they compared a baseline model that used the common method of flattening the final convolution operation and classifying through a layer of fully connected scalar neurons with a variety of configurations of homogeneous vector capsules (HVCs).  The authors proposed and demonstrated that classifying through HVCs is superior to classifying through a layer of fully connected scalar neurons on three different datasets with differing difficulty profiles.  In this paper, we extend that work to include this micro-PCB dataset.  In all experiments performed, we compare classifying through a fully connected layer of scalar neurons to the best performing HVC configuration for the simple monolithic CNN in~\cite{Byerly2019}.  In our experiments, we label the fully connected network M1 and the network using HVCs M2.

In addition to investigating the impact of HVCs on this dataset, we investigated (a) the ability of the networks to accurately predict novel rotations and perspectives of the micro-PCBs by excluding training samples with similar rotations and perspectives and (b) the ability of data augmentation techniques to mimic the excluded training samples.

\autoref{tab:experimental_design_e} shows rotations and perspectives that were used during training for experiments E1-E9.  Testing always included all images from all rotations and perspectives.  For these experiments, data augmentation techniques were \textit{not} used to simulate the rotations and perspectives that were excluded during training.  The results of those experiments is shown in \autoref{tab:experimental_results_e}.

\autoref{tab:experimental_design_a} shows, for experiments A1-A16, both which rotations and perspectives were used during training, as well as whether data augmentation techniques were used to simulate the excluded rotations, excluded perspectives, or both.  Again, testing always included all images from all rotations and perspectives.  The results of those experiments is shown in \autoref{tab:experimental_results_a}.

\autoref{tab:experimental_results_all} shows the results of a final set of experiments wherein we included all training samples and for each training sample, we applied a range of rotation and perspective warp augmentations based on the PCBs' coded labels for rotation and perspectives consistent with the distribution presented in \autoref{tab:angles_of_rotated_images} and \autoref{tab:ratios_of_perspectives}.

For all experiments, including those in which data augmentation techniques were not used \textit{to simulate the rotations and perspectives} that were excluded, a small amount of random translation was applied during training in order to encourage \textit{translational invariance}.  This translation was limited to no more than 5\% in either or both of the \(x\) and \(y\) directions.

\begin{table}[!ht]
  \centering
  \begin{minipage}{\textwidth}
    \caption{Experimental Design for Experiments Excluding Rotations and Perspectives}
    \begin{tabularx}{\textwidth}{@{}cXcccccccc@{}}
      \toprule
        & &
        \multicolumn{4}{c}{Train Rotations} &
        \multicolumn{4}{c}{Train Perspectives} \\
      \midrule
        \multirow{2}{*}{Experiment} &
        \multirow{2}{*}{} &
        Left & Left & Right & Right & Neg. & Neg. & Pos. & Pos. \\
        & & Wide & Shallow & Shallow & Wide & Far & Near & Near & Far \\
      \midrule
        E1 & & \checkmark{} & \checkmark{} & \checkmark{} & \checkmark{} & \checkmark{} & \checkmark{} & \checkmark{} & \checkmark{} \\
        E2 & & \checkmark{} & \checkmark{} & \checkmark{} & \checkmark{} & & \checkmark{} & \checkmark{} & \\
        E3 & & \checkmark{} & \checkmark{} & \checkmark{} & \checkmark{} & & & & \\
      \midrule
	      E4 & & & \checkmark{} & \checkmark{} & & \checkmark{} & \checkmark{} & \checkmark{} & \checkmark{} \\
	      E5 & & & \checkmark{} & \checkmark{} & & & \checkmark{} & \checkmark{} & \\
	      E6 & & & \checkmark{} & \checkmark{} & & & & & \\
      \midrule
        E7 & & & & & & \checkmark{} & \checkmark{} & \checkmark{} & \checkmark{} \\
	      E8 & & & & & & & \checkmark{} & \checkmark{} & \\
	      E9 & & & & & & & & & \\
      \bottomrule
    \end{tabularx}\\\label{tab:experimental_design_e}
    \captionsetup{justification=justified,singlelinecheck=false}
    \caption*{Checkmarks indicate training included images for the specified rotations or perspectives.}
  \end{minipage}
\end{table}

\begin{table}[!ht]
  \centering
  \begin{minipage}{\textwidth}
    \caption{Results of Experiments Excluding Rotations and Perspectives}
    \begin{tabularx}{\textwidth}{@{}cXrrrrrrr@{}}
      \toprule
        & &
        \multicolumn{3}{c}{M1 (Fully Connected)} &
        \multicolumn{3}{c}{M2 (Capsules)} &
        p-value \\
        \midrule
        Experiment & &
        \multicolumn{1}{c}{Mean} &
        \multicolumn{1}{c}{Max} &
        \multicolumn{1}{c}{S.D.} &
        \multicolumn{1}{c}{Mean} &
        \multicolumn{1}{c}{Max} &
        \multicolumn{1}{c}{S.D.} & \\
        \midrule
	      E1 & & 93.28\% &  96.90\% & 0.02025       & \textbf{99.45\%} & 100.00\% & 0.01117 & 0.00033        \\
	      E2 & & 92.30\% &  92.30\% &       0       & \textbf{98.84\%} &  99.14\% & 0.00244 & \num{6.68e-12} \\
	      E3 & &  9.78\% &  11.76\% & 0.01415       & \textbf{39.45\%} &  42.55\% & 0.02703 & \num{2.11e-8}  \\
        \midrule
	      E4 & & 92.30\% &  92.30\% & \num{2.58e-8} & \textbf{98.92\%} &  99.45\% & 0.00533 & \num{3.05e-9}  \\
	      E5 & & 25.14\% &  35.10\% & 0.06826       & \textbf{89.74\%} &  91.63\% & 0.02087 & \num{3.71e-8}  \\
	      E6 & & 18.10\% &  71.67\% & 0.19831       & \textbf{34.29\%} &  42.73\% & 0.05567 & 0.02304        \\
        \midrule
	      E7 & & 22.89\% &  41.01\% & 0.10565       & \textbf{78.62\%} &  81.77\% & 0.02368 & \num{2.95e-6}  \\
	      E8 & & 44.98\% & 100.00\% & 0.44777       & \textbf{46.48\%} &  50.37\% & 0.03487 & 0.91692        \\
	      E9 & & 10.10\% &  13.05\% & 0.01831       & \textbf{27.61\%} &  36.45\% & 0.05395 & 0.00013        \\
      \bottomrule
    \end{tabularx}\\\label{tab:experimental_results_e}
    \captionsetup{justification=justified,singlelinecheck=false}
    \caption*{In all cases, model M2 achieved a higher mean accuracy.  5 trials of each of experiments E1-E5, E7, and E9 were conducted.  10 trials of E6 were conducted in order to establish statistical significance.  After 10 trials of experiment E8, the higher mean of accuracy of model M2 was not shown to be statistically significant.}
  \end{minipage}
\end{table}

\begin{table}[!ht]
  \centering
  \begin{minipage}{\textwidth}
    \caption{Experimental Design for Experiments using Data Augmentation to Simulate the Excluded Rotations and Perspectives}
    \begin{tabularx}{\textwidth}{@{}cXcccccccccc@{}}
      \toprule
        & &
        \multicolumn{2}{c}{Augment the Excluded} &
        \multicolumn{4}{c}{Train Rotations} &
        \multicolumn{4}{c}{Train Perspectives} \\
      \midrule
        \multirow{2}{*}{Experiment} &
        \multirow{2}{*}{} &
        \multirow{2}{*}{Rotations} &
        \multirow{2}{*}{Perspectives} &
        Left & Left & Right & Right & Neg. & Neg. & Pos. & Pos. \\
        & & & & Wide & Shallow & Shallow & Wide & Far & Near & Near & Far \\
      \midrule
        A1 & & & \checkmark{} & \checkmark{} & \checkmark{} & \checkmark{} & \checkmark{} & & \checkmark{} & \checkmark{} & \\
        A2 & & & \checkmark{} & \checkmark{} & \checkmark{} & \checkmark{} & \checkmark{} & & & & \\
      \midrule
        A3 & & \checkmark{} & & & \checkmark{} & \checkmark{} & & \checkmark{} & \checkmark{} & \checkmark{} & \checkmark{} \\
        A4 & & \checkmark{} & \checkmark{} & & \checkmark{} & \checkmark{} & & & \checkmark{} & \checkmark{} & \\
        A5 & & \checkmark{} & & & \checkmark{} & \checkmark{} & & & \checkmark{} & \checkmark{} & \\
        A6 & & & \checkmark{} & & \checkmark{} & \checkmark{} & & & \checkmark{} & \checkmark{} & \\
        A7 & & \checkmark{} & \checkmark{} & & \checkmark{} & \checkmark{} & & & & & \\
        A8 & & \checkmark{} & & & \checkmark{} & \checkmark{} & & & & & \\
        A9 & & & \checkmark{} & & \checkmark{} & \checkmark{} & & & & & \\
      \midrule
        A10 & & \checkmark{} & & & & & & \checkmark{} & \checkmark{} & \checkmark{} & \checkmark{} \\
        A11 & & \checkmark{} & \checkmark{} & & & & & & \checkmark{} & \checkmark{} & \\
        A12 & & \checkmark{} & & & & & & & \checkmark{} & \checkmark{} & \\
        A13 & & & \checkmark{} & & & & & & \checkmark{} & \checkmark{} & \\
        A14 & & \checkmark{} & \checkmark{} & & & & & & & & \\
        A15 & & \checkmark{} & & & & & & & & & \\
        A16 & & & \checkmark{} & & & & & & & & \\
      \bottomrule
    \end{tabularx}\\\label{tab:experimental_design_a}
    \captionsetup{justification=justified,singlelinecheck=false}
    \caption*{Checkmarks in the second and third column indicate if data augmentation was used to simulate the excluded rotations, perspectives or both.  Checkmarks in the following columns indicate training included images for the specified rotations or perspectives.}
  \end{minipage}
\end{table}

\begin{table}[!ht]
  \centering
  \begin{minipage}{\textwidth}
    \caption{Results of Experiments using Data Augmentation to Simulate the Excluded Rotations and Perspectives}
    \begin{tabularx}{\textwidth}{@{}cXrrrrrrr@{}}
      \toprule
        & &
        \multicolumn{3}{c}{M1 (Fully Connected)} &
        \multicolumn{3}{c}{M2 (Capsules)} &
        p-value \\
      \midrule
        Experiment & &
        \multicolumn{1}{c}{Mean} &
        \multicolumn{1}{c}{Max} &
        \multicolumn{1}{c}{S.D.} &
        \multicolumn{1}{c}{Mean} &
        \multicolumn{1}{c}{Max} &
        \multicolumn{1}{c}{S.D.} & \\
      \midrule
         A1 & & 92.30\%          &  92.30\% &             0 & \textbf{99.27\%} &  99.69\% & 0.00357 & \num{8.44e-11} \\
         A2 & & 35.73\%          & 100.00\% &       0.39620 & \textbf{48.90\%} &  58.56\% & 0.06991 & 0.48483        \\
      \midrule
	       A3 & & 90.76\%          &  92.30\% & \num{3.44e-2} & \textbf{99.40\%} &  99.82\% & 0.00284 & 0.00052        \\
         A4 & & 22.73\%          &  26.17\% &       0.03215 & \textbf{96.37\%} &  97.48\% & 0.00785 & \num{2.95e-11} \\
	       A5 & & 23.82\%          &  31.40\% &       0.05243 & \textbf{94.52\%} &  95.75\% & 0.02143 & \num{2.93e-9}  \\
         A6 & & 20.87\%          &  29.62\% &       0.05037 & \textbf{92.51\%} &  93.97\% & 0.01614 & \num{1.53e-9}  \\
	       A7 & & \textbf{83.44\%} & 100.00\% &       0.15976 & 36.44\%          &  44.58\% & 0.06076 & 0.00027        \\
         A8 & & \textbf{58.13\%} & 100.00\% &       0.37197 & 40.87\%          &  44.95\% & 0.03404 & 0.33187        \\
         A9 & & 10.00\%          &  11.45\% &       0.00899 & \textbf{45.04\%} &  53.51\% & 0.05463 & \num{6.05e-7}  \\
      \midrule
	      A10 & & 24.14\%          &  29.43\% &       0.04385 & \textbf{93.85\%} &  94.70\% & 0.01037 & \num{5.33e-10} \\
        A11 & & \textbf{98.03\%} & 100.00\% &       0.04406 & 49.62\%          &  62.87\% & 0.10311 & \num{1.10e-5}  \\
        A12 & & \textbf{72.77\%} & 100.00\% &       0.39731 & 58.00\%          &  69.77\% & 0.08962 & 0.44100        \\
        A13 & & 23.72\%          &  51.54\% &       0.20104 & \textbf{49.68\%} &  57.27\% & 0.08771 & 0.02941        \\
        A14 & & 21.01\%          &  61.82\% &       0.22826 & \textbf{35.07\%} &  39.96\% & 0.05824 & 0.21863        \\
        A15 & & 18.88\%          &  53.63\% &       0.19435 & \textbf{34.29\%} &  41.56\% & 0.05613 & 0.12698        \\
        A16 & & \textbf{40.46\%} & 100.00\% &       0.42248 & 36.11\%          &  47.48\% & 0.06581 & 0.82585        \\
      \bottomrule
    \end{tabularx}\\\label{tab:experimental_results_a}
    \captionsetup{justification=justified,singlelinecheck=false}
    \caption*{In 11 out of 16 experiments, model M2 achieved a higher mean accuracy.  5 trials of all experiments were conducted.  In all but 4 experiments, there was greater variance across trials with model M1.}
  \end{minipage}
\end{table}

\section{Discussion}\label{sec:Discussion}

\subsection{Experiments Excluding Rotations and Perspectives During Training and Using No Data Augmentation Beyond a Small Amount of Translation --- Experiments E1-E9}

These experiments showed that model M2 (using HVCs) is superior to model M1 (using a fully connected layer) for all 9 experiments, though this was statistically significant for only 8 of the 9 experiments.  Both models M1 and M2 were better able to cope with excluded rotations than excluded perspectives.  This is not especially surprising given that rotation is an affine transformation whereas perspective changes are not.  For both M1 and M2, accuracy was especially poor when excluding all non-neutral perspectives (experiments E3, E6, and E9) to the point that the accuracy for model M1 for experiments E3 (including all rotation variants) and E9 (including no rotation variants) was equivalent to that of random guessing.  A surprising result is that model M1 for experiment E6 (which included only the near rotation variants) had a mean accuracy that was approximately twice as accurate as either E3 or E9 (for model M1).  One of the trials for model M1 experiment E6 achieved an accuracy approximately four times higher than the mean accuracy.  Indeed the standard deviation of the trials for model M1 experiment E6 was more than an order of magnitude higher than the standard deviations of either E3 or E9 (for model M1).  Most surprising for model M1 was that the maximum accuracy achieved by one trial of E8 (which included no rotation variants and only near perspective variants) was 100\%.  We do not have enough data to come to any firm conclusion regarding this anomaly but hypothesize that the loss manifold for model M1 for the training data used in E8 contains many global minima and that the less structured nature of fully connect layers relative to HVCs allows for greater exploration of the loss manifold.  The fact that E8 was the one experiment for which model M2 was not shown to be statistically significantly superior to model M1 would seem to support this.

\subsection{Experiments Excluding Rotations and Perspectives During Training and Using Data Augmentation to Synthesize the Excluded Data --- Experiments A1-A16}

These experiments showed that model M2 (using HVCs) is superior to model M1 (using a fully connected layer) for 11 out of 16 experiments, though this was statistically significant for only 8 of the 16 experiments.  Model M1 is superior twice with statistical significance.  The lack of statistical significance in 8 out of the 16 experiments is due to the high variance across trials for the the experiments using model M1.  A large contributor to the higher variances is the unusual number of trials using model M1 that achieved 100\% test accuracy.  The present theoretical understanding of non-convex optimization beyond its NP-hardness is limited to a small number of special cases (\cite{Ma2017}\cite{Hazan2016}\cite{Arora2016}\cite{Anandkumar2014}), non of which apply to modern CNNs for image classification.  As such, we can only offer conjecture based on observation regarding this phenomenon.  This conjecture being that the loss manifold of the test data has an unusually large number of local minima near the global minimum, and further that using an adaptive gradient descent method along with a fully connected layer for classification allows training to proceed in a highly exploratory manner (as opposed to an exploitative manner)~\cite{Byerly2019}.  Of the experiments that had trials which achieved 100\% test accuracy with model M1, none of them included the far perspectives in their training and only 2 out of the 6 included even the near perspectives (A11 and A12), neither of which included any rotations.  3 out of the 6 (A2, A7, and A11) fully simulated the excluded rotations and perspectives with data augmentation.  One experiment (A12) had a trial that achieved 100\% test accuracy with model M1 without attempting to simulate the excluded far perspectives.  One experiment (A8) had a trial that achieved 100\% test accuracy with model M1 without attempting to simulate either the excluded far or near perspectives.  And one experiment (A16) had a trial that achieved 100\% test accuracy with model M1 without attempting to simulate either the excluded wide or shallow rotations.

\subsection{Comparing Experiments E1-E9 with Experiments A1-A16}

\autoref{tab:experimental_results_all} shows a comparison of mean accuracies achieved by the trials of experiments E1-E9 with those of experiments A1-A16, grouping the experiments together by the rotations and perspectives that were excluded during training.  Not surprisingly, in all cases, the experiments that included data augmentation to replace some or all of the excluded samples achieved higher mean accuracy.  In 9 out of the 16 comparisons, the superiority was statistically significant.  Of those that did not produce a statistically significant difference, (1) experiments E2 and A1 each excluded only the far perspectives, (2) experiments E4 and A3 each excluded only the wide rotations, (3) experiments E8, A12, and A13 excluded all rotations and the far perspectives, and (4) experiments E9, A14, A15, and A16 excluded all rotations and all perspectives.  In 6 out of 7 these comparisons, the data augmentation was attempting to synthesize excluded perspectives.  As perspective warp is a non-affine transformation, it makes sense that synthesizing excluded perspectives with it meets with limited (but not no) success.

\begin{table}[!ht]
  \centering
  \begin{minipage}{\textwidth}
    \caption{Comparing Results with and without Augmentation of Excluded Rotations and Perspectives}
    \begin{tabularx}{\textwidth}{@{}cXrrcrrr@{}}
      \toprule
        \multirow{2}{*}{Experiment} & &
        \multicolumn{1}{c}{M1} &
        \multicolumn{1}{c}{M2} &
        \multirow{2}{*}{Experiment} &
        \multicolumn{1}{c}{M1} &
        \multicolumn{1}{c}{M2} &
        \multirow{2}{*}{p-value} \\
        & &
        \multicolumn{1}{c}{{\small (Fully Connected)}} &
        \multicolumn{1}{c}{{\small (Capsules)}} &
        &
        \multicolumn{1}{c}{{\small (Fully Connected)}} &
        \multicolumn{1}{c}{{\small (Capsules)}} & \\
      \midrule
        E2 & & 92.30\% & 98.84\% & A1 & 92.30\% & \textbf{99.27\%} & 0.056562 \\
      \midrule
        E3 & &  9.78\% & 39.45\% & A2 & 35.73\% & \textbf{48.90\%} & 0.022433 \\
      \midrule
        E4 & & 92.30\% & 98.92\% & A3 & 90.76\% & \textbf{99.40\%} & 0.112934 \\
      \midrule
        \multirow{3}{*}{E5} & & \multirow{3}{*}{25.14\%} & \multirow{3}{*}{89.74\%}
                & A4 & 22.73\% & \textbf{96.37\%} & 0.00016 \\
          & & & & A5 & 23.82\% & \textbf{94.52\%} & 0.00727 \\
          & & & & A6 & 20.87\% & \textbf{92.51\%} & 0.04678 \\
      \midrule
        \multirow{3}{*}{E6} & & \multirow{3}{*}{18.10\%} & \multirow{3}{*}{34.29\%}
                & A7 & \textbf{83.44\%} & 36.44\% & \num{2.46e-5} \\
          & & & & A8 & \textbf{58.13\%} & 40.87\% & 0.01604 \\
          & & & & A9 & 10.00\% & \textbf{45.04\%} & 0.00358 \\
      \midrule
        E7 & & 22.89\% & 78.62\% & A10 & 24.14\% & \textbf{93.85\%} & \num{1.05e-6} \\
      \midrule
        \multirow{3}{*}{E8} & & \multirow{3}{*}{44.98\%} & \multirow{3}{*}{46.48\%}
                & A11 & \textbf{98.03\%} & 49.62\% & 0.02226 \\
          & & & & A12 & \textbf{72.77\%} & 58.00\% & 0.26219 \\
          & & & & A13 & 23.72\% & \textbf{49.68\%} & 0.32177 \\
      \midrule
        \multirow{3}{*}{E9} & & \multirow{3}{*}{10.10\%} & \multirow{3}{*}{27.61\%}
                & A14 & 21.01\% & \textbf{35.07\%} & 0.31777 \\
          & & & & A15 & 18.88\% & \textbf{34.29\%} & 0.09152 \\
          & & & & A16 & \textbf{40.46\%} & 36.11\% & 0.05605 \\
      \bottomrule
    \end{tabularx}\\\label{tab:experiment_comparison}
    \captionsetup{justification=justified,singlelinecheck=false}
    \caption*{Horizontal lines in this table are used to group together experiments E1-E9 with their counterpart experiments A1-A16 based on the samples that were used during training.  For example, E5 and A4-A6 were all trained on the subset of training samples that excluded the wide rotations and the far perspectives.}
  \end{minipage}
\end{table}

\subsection{Regarding Synthesizing Alternate Perspectives}

\begin{figure}[!ht]
  \centering
  \begin{minipage}{.7\textwidth}
    \begin{tabularx}{\textwidth}{@{}cXc@{}}
      \includegraphics[width=2in]{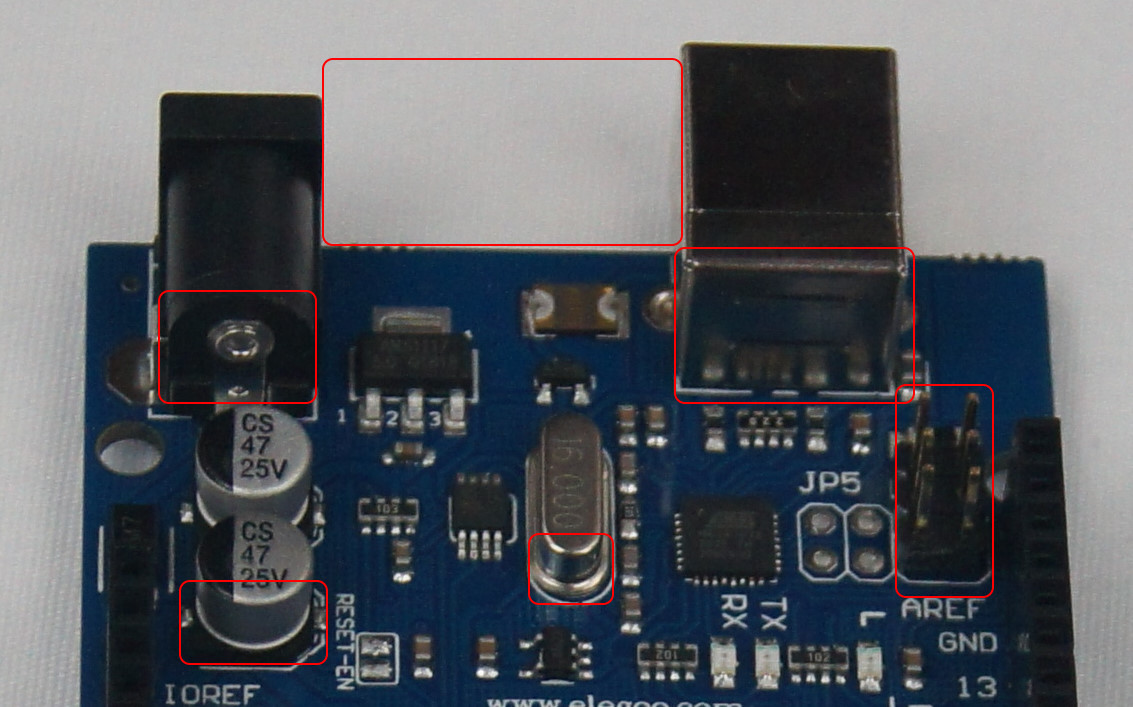} &
      &
      \includegraphics[width=2in]{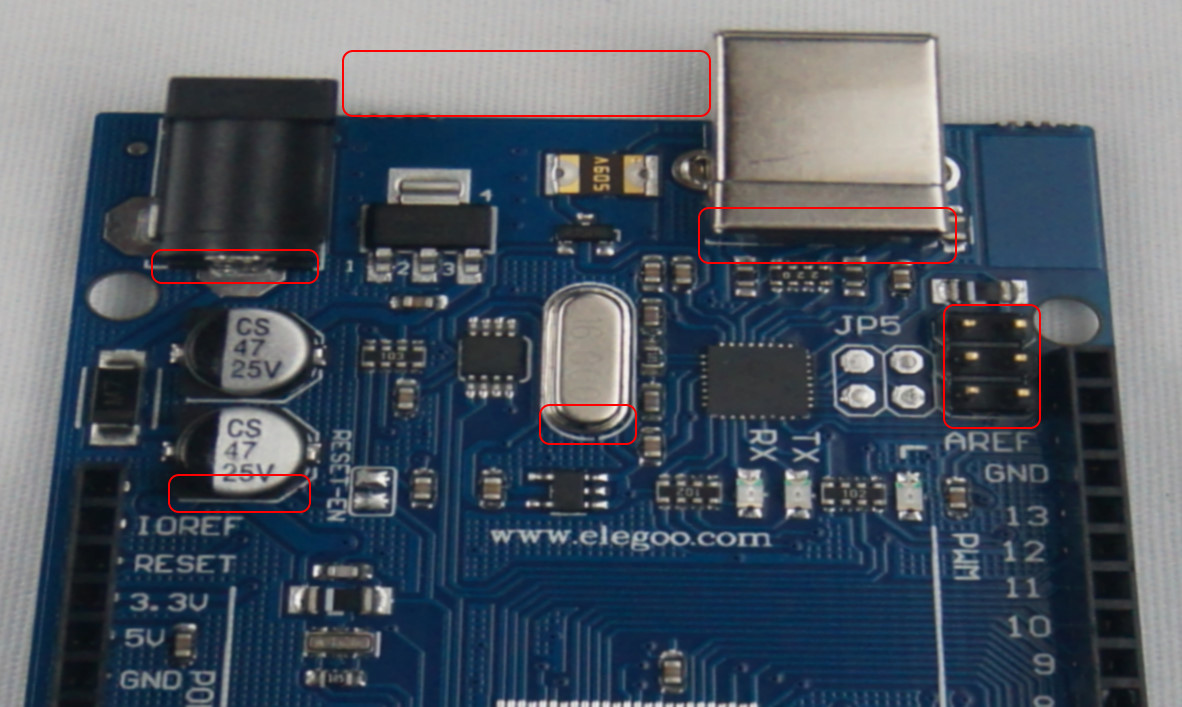} \\
    \end{tabularx}
    \caption{The image on the left shows a micro-PCB actually captured in a negative far position.  The image to the right shows the same micro-PCB captured from a neutral position and then subjected to an (exaggerated) perspective warp.  The annotating boxes show important differences between an image actually captured from a perspective position vs.\ simulating a perspective.}\label{fig:perspective_actual_vs_simulated}
  \end{minipage}
\end{figure}

As our experiments demonstrate, using data augmentation to supply excluded and/or greater variations of rotations works better than data augmentation to supply excluded and/or greater variations of perspective.  Indeed, rotation is generally considered a staple data augmentation technique irrespective of the subject matter.  As mentioned earlier, this is because rotation is an affine transformation and as such, rotating the image produces the same result as rotating the capturing apparatus would have.  Perspective warp of captured \textit{3-D} subject matter is not affine.  Moving the capturing apparatus to a different perspective relative to 3-D subject matter will produce larger or smaller patches of components that extend into the third dimension (see \autoref{fig:perspective_actual_vs_simulated}).  While our simulation of perspective differences did improve upon accuracy when not using such, our results show that, when possible, capturing a variety of perspectives during training is the best avenue for generating higher accuracy during subsequent evaluations that could include analogous perspective varieties.

\subsection{Experiments Including All Rotations and Perspectives During Training and Using Data Augmentation Augmentation to Synthesize Variations of Those Rotations and Perspectives}

For our final set of experiments, we both trained with the full training set and applied rotational and perspective warp data augmentation throughout training.  For each training sample, we rotated it by a random rotation drawn from a normal distribution with a zero mean and the standard deviation for its rotational label (see \autoref{tab:angles_of_rotated_images}).  For example, if the training sample was labeled as left wide, it was rotated by a random amount chosen from a normal distribution with a mean of 0 degrees and a standard deviation of 4.50481 degrees.  Perspective warp transformations were applied using the same procedure with standard deviations for its perspective label (see \autoref{tab:ratios_of_perspectives}).

As is shown in \autoref{tab:experimental_results_all}, model M1, using a fully connected layer, achieved a mean accuracy of 94.52\%, surpassing 15 out of 16 of the experiments detailed in \autoref{tab:experimental_results_a} (for model M1).  Experiment A11 performed slightly better, but it should be noted that that experiment also used data augmentation for both rotation and perspective warp, with the difference being that experiment A11 excluded all of the rotated training samples, and the far perspectives.  Model M2, using HVCs, achieved a mean accuracy of 99.06\%, surpassing 14 out of 16 of the experiments detailed in \autoref{tab:experimental_results_a} (for model M2).  Experiment A1 and A3 performed slightly better, but once again, it should be noted that those experiments also used data augmentation that covered the rotations and perspectives of the training samples that were excluded.  Model M2's maximum accuracy was 100\% surpassing the maximum accuracy of all experiments (for model M2) A1-A11.

\begin{table}[!ht]
  \centering
  \begin{minipage}{\textwidth}
    \caption{Results of Experiments with no Exclusions and Using Data Augmentation}
    \begin{tabularx}{\textwidth}{@{}Xrrr@{}}
      \toprule
        Model &
        \multicolumn{1}{c}{Mean} &
        \multicolumn{1}{c}{Max} &
        \multicolumn{1}{c}{S.D.} \\
        \midrule
	      M1 (Fully Connected) & 94.52\% &  96.90\% & 0.02009 \\
	      M2 (Capsules)        & \textbf{99.06\%} & \textbf{100.00\%} & 0.01862 \\
      \bottomrule
    \end{tabularx}\\\label{tab:experimental_results_all}
    \captionsetup{justification=justified,singlelinecheck=false}
    \caption*{We conducted 10 trials for each model.  M2, using homogeneous vector capsules, is shown to be superior with a p-value of \num{5.52e-5}.}
  \end{minipage}
\end{table}

\section{Conclusion}\label{sec:conclusion}

In this paper, we have presented a dataset consisting of high-resolution images of 13 micro-PCBs captured in various rotations and perspectives relative to the camera, with each sample labeled for PCB type, rotation category, and perspective category.  We have shown, experimentally, that (1) classification of these micro-PCBs from novel rotations and perspectives is possible, but, in terms of perspectives, better accuracy is achieved when networks have been trained on representative examples of the perspectives that will be evaluated.  (2) That, even though perspective warp is non-affine, using it as a data augmentation technique in the absence of training samples from actually different perspectives is still effective and improves accuracy.  (3) And that using homogeneous vector capsules (HVCs) is superior to using fully connected layers in convolutional neural networks, especially when the subject matter has many sub-components that vary equivariantly (as is the case with micro-PCBs), and when using the full training dataset and applying rotational and perspective warp data augmentation the mean accuracy of the network using HVCs is 4.8\% more accurate then when using a fully connected layer for classification.

\printbibliography{}

\end{document}